\documentclass{article}

\newif\ifnotes
\notestrue

\usepackage{microtype}
\usepackage{graphicx}
\usepackage{booktabs} 
\usepackage{caption}
\usepackage{subcaption}
\usepackage{hyperref}

\usepackage[dvipsnames]{xcolor}
\usepackage{listings}
\usepackage{float}
\usepackage{marginnote}
\usepackage{tabularx}
\usepackage{ragged2e}

\definecolor{codegreen}{rgb}{0,0.6,0}
\definecolor{codegray}{rgb}{0.5,0.5,0.5}
\definecolor{codepurple}{rgb}{0.58,0,0.82}
\definecolor{backcolour}{rgb}{0.95,0.95,0.92}

\lstdefinestyle{mystyle}{
  backgroundcolor=\color{backcolour},
  commentstyle=\color{codegreen},
  keywordstyle=\color{magenta},
  numberstyle=\tiny\color{codegray},
  stringstyle=\color{codepurple},
  basicstyle=\ttfamily\small,
  breakatwhitespace=false,
  breaklines=true,
  captionpos=b,
  keepspaces=true,
  numbers=left,
  numbersep=5pt,
  showspaces=false,
  showstringspaces=false,
  showtabs=false,
  tabsize=2 }
\newfloat{lstfloat}{tbp}{lop}
\floatname{lstfloat}{Listing}

\usepackage[firstpage]{draftwatermark}
\SetWatermarkText{
 \hspace*{4.5in}
 \raisebox{10in}{
  \includegraphics[height=0.9in]{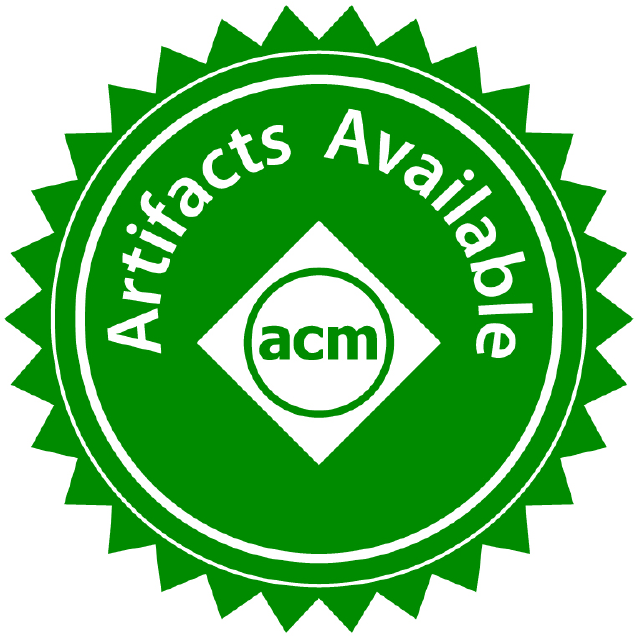}
  \hspace{-1in}
  \includegraphics[height=0.9in]{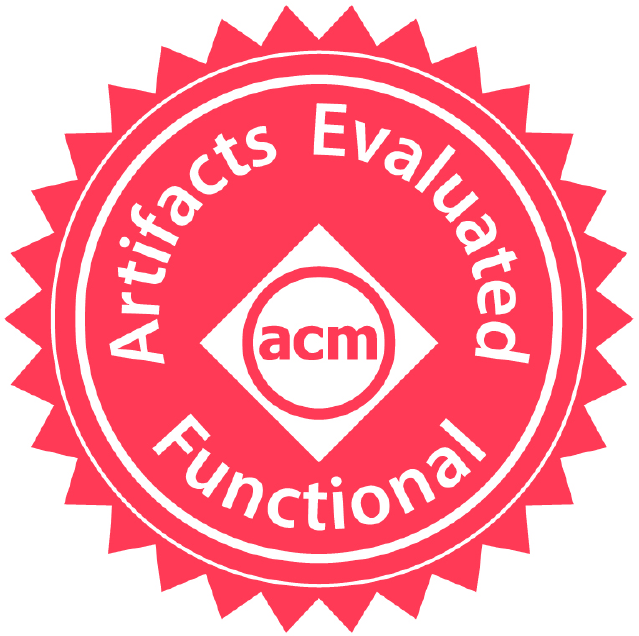}
 }
}
\SetWatermarkAngle{0}

\usepackage[accepted]{mlsys2023}

\usepackage{amsmath,amsthm,amsfonts,bm,mathtools}
\usepackage{tabularx,threeparttable,array,colortbl}
\usepackage{xspace}
\usepackage{calc}
\usepackage[inline]{enumitem}
\usepackage[title]{appendix}
\usepackage{dblfloatfix}
\usepackage{cleveref}

\newcommand{\review}[1]{#1}         

\newcommand{\oursystem}         {\review{SALIENT++}\xspace}

\newcommand{\magcites}{\textit{mag240c}\xspace}
\newcommand{\magcitesfigs}{mag240c\xspace}

\newcommand{\defn}[1]       {{\textit{\textbf{#1}}}}
\newcommand{\ang}[1]            {\ifmmode{\mathopen{}\left\langle #1 \right\rangle\mathclose{}}
                                 \else{$\mathopen{}\left\langle${#1}$\right\rangle\mathclose{}$}\fi}

\newtheorem{theorem}            {Theorem}

\newtheorem{proposition}[theorem]{Proposition}

\newcommand{\punt}[1]{}

\newcommand{\secref}[1]         {Section~\ref{sec:#1}}

\newcommand{\figlabel}[1]   {\label{fig:#1}}

\newcommand{\seclabel}[1]   {\label{sec:#1}}

\newcommand{\figref}[1]         {Figure~\ref{fig:#1}}

\newcommand{\tabref}[1]         {Table~\ref{tab:#1}}

\newcounter{ccount}

\newcommand*\makeAlph[1]{%
	  \ifnum#1<1\else
	      \ifnum#1>26 a\makeAlph{\numexpr#1-26}
	          \else\symbol{\numexpr96+#1}\fi\fi}

\newcounter{ccountletters}

\MakeRobust{\Call}
\algnewcommand{\LineComment}[1]{\State \(\triangleright\) #1}
\newcolumntype{Y}{>{\centering\arraybackslash}X}

\crefname{section}{Section}{Sections}
\crefname{appendix}{Appendix}{Appendices}
\crefname{figure}{Figure}{Figures}
\crefname{table}{Table}{Tables}
\crefname{equation}{equation}{equations}
\Crefname{equation}{Equation}{Equations}
\crefname{theorem}{Theorem}{Theorems}
\crefname{lemma}{Lemma}{Lemmas}
\crefname{corollary}{Corollary}{Corollaries}
\crefname{definition}{Definition}{Definitions}\crefname{proposition}{Proposition}{Propositions}
\crefname{remark}{Remark}{Remarks}

\newcommand{\graph}                 {G}
\newcommand{\vertices}              {\set{V}}
\newcommand{\edges}                 {\set{E}}
\newcommand{\trainvertices}         {\set{T}}
\newcommand{\numvertices}           {N}
\newcommand{\numedges}              {M}
\newcommand{\weight}                {t}

\newcommand{\nhood}                 {\set{N}}
\newcommand{\nhoodsampled}          {\nhood^{s}}
\renewcommand{\deg}                 {d}
\newcommand{\fanout}                {f}
\newcommand{\layer}                 {\ell}
\newcommand{\numlayers}             {L}
\newcommand{\hop}                   {h}
\newcommand{\batch}                 {\set{B}}
\newcommand{\batchsize}             {B}
\renewcommand{\part}                {k}
\newcommand{\numparts}              {K}
\newcommand{\oflayer}[1][\layer]    {_{#1}}
\newcommand{\ofhop}[1][\hop]        {^{[#1]}}
\newcommand{\ofpart}[1][\part]      {_{#1}}
\newcommand{\numdim}                {D}

\newcommand{\repfactor}             {\alpha}

\newcommand{\gpufactor}             {\beta}

\newcommand{\set}[1]                {\mathcal{\MakeUppercase{#1}}}

\newcommand{\union}                 {\mathbin{\cup}}
\newcommand{\intersection}          {\mathbin{\cap}}
\newcommand{\bigunion}              {\bigcup}
\newcommand{\bigintersection}       {\bigcap}

\DeclareMathOperator*{\prob}        {Pr}

\newcommand{\Prob}[1]               {\prob \left[ #1 \right]}

\DeclarePairedDelimiter{\abs}       {\lvert}{\rvert}

\DeclarePairedDelimiter{\setenum}   {\lbrace}{\rbrace}
\DeclarePairedDelimiterX{\setcond}[2]%
    {\lbrace}{\rbrace}{#1 \; \delimsize\vert \; \mathopen{} #2}
\DeclarePairedDelimiterX{\probcond}[2]%
    {[}{]}{#1 \; \delimsize\vert \; \mathopen{} #2}
\newcommand*{\setcard}              {\abs}

\newcommand{\speedup}[1]            {$#1\times$}

\newcommand{\tp}                    {\mathsf{T}}

\let\leftorig\left
\let\rightorig\right
\renewcommand{\left}{\mathopen{}\mathclose\bgroup\leftorig}
\renewcommand{\right}{\aftergroup\egroup\rightorig}

\graphicspath{{figs/}{figs_matlab/}}

\newcommand{\paperdata}[1]{#1}

\newcommand{\real}{\mathbb{R}}

\newcommand{\tfknote}[1]{}
\newcommand{\tbsnote}[1]{}
\newcommand{\asinote}[1]{}
\newcommand{\jcnote}[1]{}
\newcommand{\philnote}[1]{}

\newcommand{\clearpagedraft}{}

\definecolor{Gray}{gray}{0.9}

\algnewcommand\algorithmicforeach{\textbf{for each}}
\algdef{S}[FOR]{ForEach}[1]{\algorithmicforeach\ #1\ \algorithmicdo}

\makeatletter
\newcommand{\multlinealgo}[1]{%
  \begin{tabularx}{\dimexpr\linewidth-\ALG@thistlm}[t]{@{}X@{}}
    #1
  \end{tabularx}
}
\makeatother

\newcolumntype{L}{>{\raggedright\arraybackslash}X}
\newcolumntype{R}{>{\raggedleft\arraybackslash}X}
\newcolumntype{C}{>{\centering\arraybackslash}X}

\mlsystitlerunning{Communication-Efficient Graph Neural Networks with
  Probabilistic Neighborhood Expansion Analysis and Caching}

\begin{document}

\twocolumn[
\mlsystitle{Communication-Efficient Graph Neural Networks with Probabilistic
  Neighborhood Expansion Analysis and Caching}



\mlsyssetsymbol{equal}{*}

\begin{mlsysauthorlist}
\mlsysauthor{Tim Kaler}{equal,mit,mitibm}
\mlsysauthor{Alexandros-Stavros Iliopoulos}{equal,mit,mitibm}
\mlsysauthor{Philip Murzynowski}{equal,mit,mitibm}
\mlsysauthor{Tao B. Schardl}{mit,mitibm}
\mlsysauthor{Charles E. Leiserson}{mit,mitibm}
\mlsysauthor{Jie Chen}{mitibm,ibm}
\end{mlsysauthorlist}

\mlsysaffiliation{mit}{MIT CSAIL}
\mlsysaffiliation{mitibm}{MIT-IBM Watson AI Lab}
\mlsysaffiliation{ibm}{IBM Research}

\mlsyscorrespondingauthor{Jie Chen}{chenjie@us.ibm.com}

%
%
%
%
%

\mlsyskeywords{Graph neural network, distributed multi-GPU training, neighborhood expansion, caching}

\vskip 0.3in

\begin{abstract}


Training and inference with graph neural networks (GNNs) on massive graphs has been actively studied since the inception of GNNs, owing to the widespread use and success of GNNs in applications such as recommendation systems and financial forensics. This paper is concerned with minibatch training and inference with GNNs that employ node-wise sampling in distributed settings, where the necessary partitioning of vertex features across distributed storage causes feature communication to become a major bottleneck that hampers scalability.
To significantly reduce the communication volume without compromising prediction accuracy, we propose a policy for caching data associated with frequently accessed vertices in remote partitions.  The proposed policy is based on an analysis of vertex-wise inclusion probabilities~(VIP) during multi-hop neighborhood sampling, which may expand the neighborhood far beyond the partition boundaries of the graph.  VIP analysis not only enables the elimination of the communication bottleneck, but it also offers a means to organize in-memory data by prioritizing GPU storage for the most frequently accessed vertex features.
%
%
We present \oursystem, which extends the prior state-of-the-art SALIENT system to work with partitioned feature data and leverages the VIP-driven caching policy.  \oursystem retains the local training efficiency and scalability of SALIENT by using a deep pipeline and drastically reducing communication volume while consuming only a fraction of the storage required by SALIENT.  We provide experimental results with the Open Graph Benchmark data sets and demonstrate that training a 3-layer GraphSAGE model with \oursystem on 8 single-GPU machines is \paperdata{\speedup{7.1}} faster than with SALIENT on 1 single-GPU machine, and \paperdata{\speedup{12.7}} faster than with DistDGL on 8 single-GPU machines.
\end{abstract}
]



\printAffiliationsAndNotice{\mlsysEqualContribution} 

\section{Introduction}
\seclabel{introduction}

Graph neural networks (GNNs) are an important class of machine learning models that incorporate relational inductive bias for effective representation learning on graph structured data~\citep{Li2016, Kipf2017, Hamilton2017, Velickovic2018, Xu2019}. These neural networks have been successfully applied in a number of use cases, including product recommendation, traffic forecasting, and financial forensics~\citep{Ying2018, Li2018, Weber2019}. In many applications, data are continuously collected and the resulting graph grows rapidly, calling for efficient GNN training and inference systems that can scale with the explosive increase of graph data.

This work considers minibatch training and inference with neighborhood sampling in the distributed setting, where vertex data (features) are partitioned across machines.  As opposed to full-batch optimization, minibatch optimization is typical for training neural networks, but it poses a unique challenge for GNNs because of the exponential increase of neighborhood size across network layers~\citep{Chen2018}.  Neighborhood sampling is a popular and effective approach to mitigating this issue~\citep{Hamilton2017, Chen2018, Ying2018, Zou2019, Zeng2020, Ramezani2020, Dong2021}.  In distributed GNN training, each machine computes the training loss for a minibatch of vertices that are local to the machine's partition.  At every step of the training optimization, a set of partition-wise minibatches forms a ``distributed minibatch'' which is used to update the GNN model parameters.  Distributed GNN inference is organized similarly.

Even with neighborhood sampling, minibatch neighborhood expansion creates a communication bottleneck as each machine needs to access remote data for sampled vertices.
The communication pattern is stochastic because of the random nature of minibatch and neighborhood sampling.  
Communication time often dominates the time for training computations.  An example of this bottleneck effect is shown in \cref{tab:performance-progression}, which we will walk through shortly.

To overcome the communication bottleneck, we propose an analysis of neighborhood access patterns via vertex inclusion probabilities, as well as a caching policy based on the analysis.  We refer to this analysis as ``VIP analysis.''  In contrast to commonly used heuristics for estimating vertex access probabilities --- e.g., based on vertex degree and expanded boundary frontiers~\citep{PaGraph}, random walks~\citep{Dong2021, Min2021}, or simulated GNN computations~\citep{Yang2022} --- VIP analysis estimates access probabilities for all graph vertices based on an analytical model of the actual stochastic neighborhood expansion process.
We derive the specific VIP model for the important class of node-wise sampling schemes~\citep{Hamilton2017, Ying2018, Velickovic2018, Xu2019, liuBanditSamplers2020} and find that the resulting caching policy drastically reduces the communication volume in distributed GNNs that employ node-wise sampling with only a modest memory overhead.  Moreover, VIP analysis offers a means to organize in-memory data by prioritizing GPU storage for the most frequently accessed vertex features, reducing host-to-device data transfers.

Our system, \oursystem, is developed over SALIENT \citep{salient}, an efficient GNN system that achieves state-of-the-art performance through performance-engineered neighborhood sampling, shared-memory parallel batch preparation, and data-transfer pipelining. SALIENT achieves per-epoch time that is nearly equal to the GPU time for model computation alone (effectively hiding the cost of minibatch construction, data transfer, and gradient communication), and it scales well with additional machines.  But SALIENT has the drawback of replicating the entire data set on each machine, imposing a hard limit on the data set size.  Extending SALIENT to distribute vertex feature data requires addressing the feature communication bottleneck.

\begin{figure*}
    %
    \begin{minipage}[c]{0.640\linewidth}
    \centering
    \includegraphics[width=\linewidth]{%
        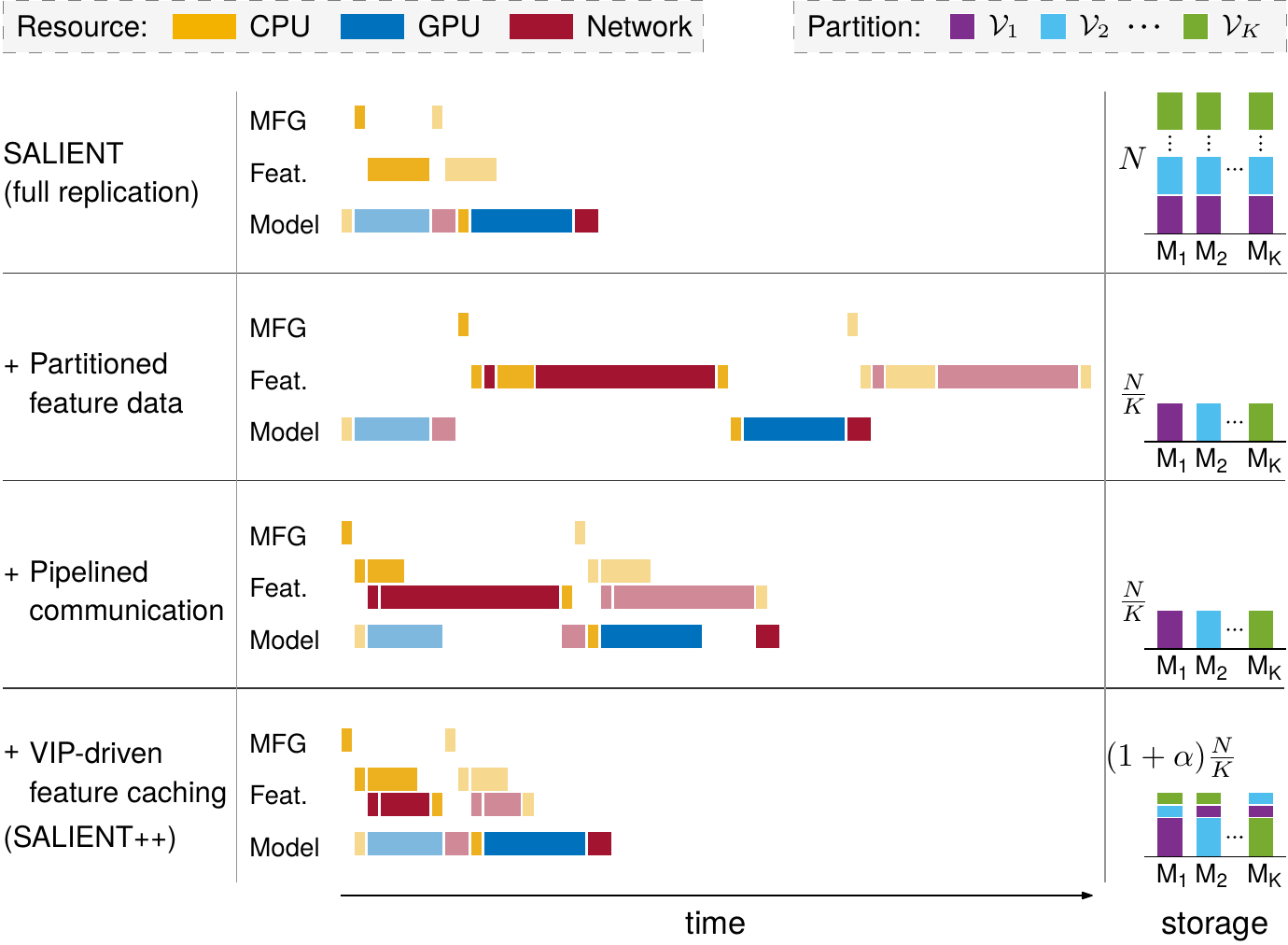}
    \end{minipage}%
    %
    \hspace*{\fill}
    \begin{minipage}[c]{0.320\linewidth}
    \vspace*{-2ex}
    \caption{Illustration of distributed GNN computation profiles and feature data storage for the precursor system SALIENT and successive optimizations leading to our system \oursystem.  
    \textbf{Left:}~Utilization of system resources (CPU, GPU, and network) in one machine.  The processing of each minibatch is broken into three parts. ``MFG'':~message-flow graph construction. ``Feat.'':~local feature tensor slicing, feature communication and joining, and host-to-device transfers. ``Model'':~forward and backward passes, plus communication among machines for model updates.  In addition to the highlighted minibatch, the profiles also depict parts of a preceding and a succeeding minibatch in faded colors to show overlap between pipelined operations.
    \textbf{Right:}~The color-stacked bars indicate the number of locally stored feature vectors in each of $K$ total machines.}
    \label{fig:overview}
    \end{minipage}
\end{figure*}

\oursystem uses the VIP-analysis-driven caching policy and a deep pipeline to achieve scalability and efficiency. It nearly matches the performance of SALIENT with only a fraction of SALIENT's memory requirements. To highlight the efficacy of \oursystem, \cref{tab:performance-progression} lists the resulting performance of progressive modifications on top of SALIENT.  This example uses the ogbn-papers100M data set~\citep{Hu2020} and partitions the graph using METIS~\citep{metis} with an edge-cut minimization objective and balancing constraints for the number of training, validation, and overall vertices, as well as the total number of edges, in each partition.  Accompanying \cref{tab:performance-progression}, \cref{fig:overview} illustrates the key differences between SALIENT and the successive optimizations in \oursystem using simplified computation profiles and storage requirement depictions.

\begin{table}
  \caption{Per-epoch runtime of a progressively more sophisticated distributed GNN
    training system on the undirected ogbn-papers100M data set, using a 3-layer
    GraphSAGE architecture with sampling fanouts (15,10,5) and a hidden
    layer dimension of 256.  For the system with remote feature caching, the size of the cache relative to the size of local-vertex features for each machine was 8\% (2 machines), 16\% (4 machines), or 32\% (8 machines).}
  \label{tab:performance-progression}
  \vskip 0.05in
  \centering
  \small
  \begin{tabularx}{\linewidth}{@{} Xrrrr @{}}
    \toprule
    
    & \multicolumn{4}{c@{}}{\emph{No.\ of machines}} \\
    \cmidrule(l){2-5}
    \emph{System} & 1     & 2      & 4       & 8 \\
    \midrule
    \emph{SALIENT (Full replication)}       & 20.7s & 10.76s &  6.02s  &  3.08s\\
    \emph{ + Partitioned features}              & ---   & 33.04s & 15.98s  & 10.85s\\
    \emph{ + Pipeline communication}        & ---   & 16.12s &  8.73s  & 5.43s\\
    \emph{ + Feature caching}          & ---   & 10.51s &  5.45s  &  2.91s\\
    \bottomrule
  \end{tabularx}
\end{table}

We first explore the distributed performance of SALIENT with a disjoint partitioning of feature data across machines.  
We modify SALIENT to only replicate the graph structure and communicate vertex features on demand and in bulk for each minibatch and its sampled neighborhood.  Although distributing the vertex features reduces SALIENT's memory requirement substantially, it also leads to an immediate performance degradation.  Specifically, we observe slowdown of roughly \paperdata{$2$--$3\times$} on two or four machines and \paperdata{$3.5\times$} on eight machines relative to SALIENT with full-replication.  This slowdown presents itself in the form of high latencies between GNN model computations (row 2 in \cref{fig:overview}).  Overlapping feature communication with other GNN operations (sampling, local feature slicing, host-to-device transfers, training computations, and model updates) through pipelining offers some improvement, but the system remains bottlenecked by communication (row 3 in \cref{fig:overview}).

If we allow a small memory overhead to cache frequently accessed remote features locally, however, the combined effects of reduced communication volume and pipelining boost performance substantially.  The sheer reduction of communication volume due to caching makes it possible for pipelining to (nearly) fully hide communication (4th row in \cref{fig:overview}).  On $K$ machines, \oursystem reduces the total feature memory footprint of SALIENT by almost a factor of $K$ while matching the performance achieved by SALIENT's full-replication strategy.

In summary, this work makes the following contributions:
\begin{enumerate}[leftmargin=*,topsep=0pt]
  \setlength\itemsep{0em}

  \item An analysis of vertex inclusion probabilities~(VIP) during GNN neighborhood expansion with node-wise sampling.  We show how VIP analysis can be used to minimize
        the expected communication among machines as well as the host-to-device data transfers within each machine.

  \item The design and implementation of \oursystem, a system for distributed
        GNN training and inference that is both efficient and scalable.
        The system is based on two key components: a maximum-likelihood static
        caching policy for remote and local features based on VIP analysis; and a deep pipeline that overlaps feature communication with
        other GNN operations, ensuring high utilization of available network
        bandwidth.


	%
	
  \item A systematic evaluation of the end-to-end training performance and
	  scalability of \oursystem on three large benchmark graphs. On the largest of these
		data sets, which does not fit in-memory on one machine,
		\oursystem achieves a speedup of \paperdata{\speedup{1.75}} when scaling from 4
		to 8 single-GPU machines and an additional \paperdata{\speedup{1.45}} when scaling from 8 to 16 single-GPU machines.

\end{enumerate}

    \clearpagedraft
\section{Background and related work}
\seclabel{background}

This section briefly recalls background and introduces notation for GNNs and their training/inference. It also discusses related work on systems and distributed training.

\subsection{Graph neural networks}
This work focuses on the class of \emph{message passing neural networks} (MPNNs)~\citep{Gilmer2017}, which encompasses many GNNs of major interest for massive graphs. Let $\graph=(\vertices,\edges)$ be a graph with $\vertices$ being the vertex set and $\edges$ being the edge set. Let $X\in\real^{\numvertices\times \numdim}$ be the vertex feature matrix, where $\numvertices$ is the number of vertices and $\numdim$ is the feature dimension. A row of $X$, denoted by $x_v\in\real^\numdim$, is the feature vector of a vertex $v$. Let $\ell=0,\ldots,\numlayers$ denote the layer index and $\nhood_{1}(v)$ be the one-hop neighborhood of $v$. MPNNs use the following update rule to define a layer:
\begin{equation}
\label{eqn:MPNN}
h_v^{\ell} = 
\textsc{upd}^{\ell}\Big( h_v^{\ell-1}, 
\textsc{agg}^{\ell}\big( \{ h_u^{\ell-1} \mid u \in \nhood_{1}(v) \} \big) \Big) ,
\end{equation}
where $h_v^{\ell}$ is the layer-$\ell$ representation of $v$,
$\textsc{agg}^{\ell}$ is a set aggregation function, $\textsc{upd}^{\ell}$ is a
two-argument update function, and $h_v^0=x_v$. One sees that the
representation of a vertex $v$ depends on the past-layer representations of $v$ and its neighbors in $\mathcal{N}_1(v)$; and this dependence is recursive.

GNNs differ in their design of the two functions
in~\eqref{eqn:MPNN}. For example, in GraphSAGE~\citep{Hamilton2017},
$\textsc{agg}^{\ell}$ is a mean, LSTM, or pooling operator, and
$\textsc{upd}^{\ell}$ concatenates the two arguments and applies a
linear layer.  In GIN~\citep{Xu2019}, $\textsc{agg}^{\ell}$ is the sum
of $\{h_u^{\ell-1}\}$ and $\textsc{upd}^{\ell}$ is the sum of its
arguments followed by an MLP\@.  In GAT~\citep{Velickovic2018},
$\textsc{agg}^{\ell}$ is the identity and $\textsc{upd}^{\ell}$
computes $h_v^{\ell}$ as a weighted combination of
$W^{\ell-1}h_u^{\ell-1}$ for all $u\in\{v\}\cup\mathcal{N}_1(v)$, where
the weights are attention coefficients and $W^{\ell-1}$ is the
parameter matrix of the layer.

\subsection{Minibatch training}
Neural networks can be trained with gradient descent (full-batch training) or stochastic gradient descent methods (minibatch training). A majority of the deep learning literature advocates minibatch training, with theoretical and empirical evidence suggesting that it converges faster, generalizes better, and is more suitable for GPU-oriented computing infrastructures~\citep{Bottou2018}.

Minibatch training incurs a unique challenge for GNNs, often termed the ``neighborhood explosion'' problem: to compute the training loss for a vertex $v$, \cref{eqn:MPNN} indicates that it requests information from the neighborhood recursively, growing its size exponentially with the number of layers. For a reasonably sized minibatch, the multi-hop neighborhood quickly covers a large portion of the graph, incurring a prohibitive time cost that eclipses the saving in convergence speed. Moreover, the storage requirement easily exceeds the memory capacity of a GPU\@.

\subsection{Neighborhood sampling}
Restricting the neighborhood size by sampling is an effective method for tackling the neighborhood explosion problem. Three major types of sampling strategies are node-wise sampling~\citep{Hamilton2017, Ying2018}, layer-wise sampling~\citep{Chen2018, Zou2019}, and subgraph sampling~\citep{Chiang2019, Zeng2020}.

Node-wise sampling approaches modify the neighborhood for each vertex $v$ separately by taking a random subset containing at most $\fanout$ neighbors (called the \emph{fanout}). Layer-wise sampling approaches collect the neighbors of all vertices in a minibatch and sample the neighborhood from their union. Such sampling proceeds recursively layer by layer and can use a nontrivial sampling distribution to control pre-activation variance while preserving unbiasedness. Nonlinear activation functions destroy unbiasedness anyway, but training convergence results can still be established based on asymptotic consistency~\citep{Chen2018a}. Subgraph sampling approaches sample a connected subgraph and compute the minibatch loss restricted to this subgraph. This work focuses on node-wise sampling for its widespread use.

\subsection{Minibatch inference}
Similar to training, inference can be performed in either full batches or minibatches. Whereas the choice between full batch or minibatch training centers on convergence and generalization,
the choice made for inference depends on the computational pattern and implementation effort. Full-batch inference avoids neighborhood sampling but requires a separate implementation of the forward pass for different GNN architectures. On the other hand, minibatch inference can reuse the forward function for training, improving productivity for application developers and architecture designers, but the stochastic nature of neighborhood sampling may return different predictions and does not guarantee the same accuracy as the case of not performing sampling. Nevertheless, the authors of SALIENT demonstrate strong empirical results that show that prediction accuracy is not compromised with a reasonable choice of the fanouts~\citep{salient}. This work follows the practice of minibatch inference.

\subsection{Related work}
Due to the computational pattern of neighborhood aggregation, GNN training systems on massive graphs differ substantially from those for usual neural networks in design and implementation. Many GNN systems are developed based on full-batch training for simplicity, which avoids the complication of neighborhood sampling. Examples include NeuGraph~\citep{Ma2019}, Roc~\citep{Jia2020}, DeepGalois~\citep{Hoang2021}, FlexGraph~\citep{Wang2021}, Seastar~\citep{Wu2021}, GNNAdvisor~\citep{Wang2021a}, DistGNN~\citep{md2021distgnn}, and BNS-GCN~\citep{BNSGCN}. Some of these systems are built on common deep learning frameworks, such as TensorFlow~\citep{Abadi2015} and PyTorch~\citep{Paszke2019}, while others are built on self-developed programming models.

Examples of systems that perform minibatch training include DistDGL~\citep{Zheng2020}, Zero-Copy~\citep{Min2021}, GNS~\citep{Dong2021}, and $P^3$~\citep{Gandhi2021}. However, these publications report results on either multiple machines with only CPUs or a single machine with one or multiple GPUs. A newer version of DistDGL (DistDGLv2~\citep{Zheng2021}), SALIENT~\citep{salient}, and our system \oursystem 
 demonstrate results in the distributed multi-GPU setting.

\oursystem employs edge-cut graph partitioning to distribute data across machines. An alternative approach, adopted by DistGNN~\citep{md2021distgnn}, is vertex-cut partitioning, which ensures each edge is local to some machine.  A drawback of this approach
is that a vertex may be assigned to multiple machines, leading to memory overhead due to feature replication that is reported to be as high as 5$\times$.  \oursystem, on the other hand, achieves high distributed performance with less than 50\% memory overhead due to caching in our experiments.

\oursystem employs a caching policy to reduce communication and data transfer. We introduce a policy based on an analysis of vertex access probabilities during neighborhood sampling. Although caching has been explored by several other systems in this context, previous caching policies are heuristic, using proxy measures such as vertex degrees~\citep{PaGraph}, random walks~\citep{Dong2021, Min2021}, or simulated access frequencies~\citep{Yang2022}.
A complementary approach to reducing communication time is taken by DGCL~\citep{DGCL}, which uses a communication-planning algorithm to optimize the communication pattern for a specific network topology.

Marius and MariusGNN~\citep{marius, Marius++} are out-of-core
training systems that work on a single machine by exploiting external
memory. They operate in a different setting from ours, but our vertex inclusion
probability analysis may be used in complement to optimize on-disk data representation and disk I/O.
      \clearpagedraft
\section{Vertex inclusion probabilities for GNNs with node-wise sampling}
\label{sec:neighbor-access}

This section describes a principled approach to estimating and reducing data access costs in distributed GNNs with node-wise sampling.  Node-wise sampling was notably introduced in GraphSAGE~\citep{Hamilton2017} and is used with a variety of GNN architectures~\citep{Ying2018, Velickovic2018, Xu2019, liuBanditSamplers2020}.  We first introduce \defn{vertex-inclusion probability} (\defn{VIP}) analysis to calculate the probability that any vertex will be present in the sampled $\numlayers$-hop expanded neighborhood of a minibatch in some partition.  VIP analysis takes into account the distinctive structure of neighbor accesses in GNNs with minibatches and node-wise sampling.  We then demonstrate that VIP analysis enables a simple but highly effective strategy for reducing data movement during GNN computations.

The VIP model for node-wise sampling derived in this section does not apply to other sampling
schemes, such as layer-wise or subgraph sampling. It may be possible
to derive VIP models for these other sampling schemes in a similar way, but this is outside our scope.


\subsection{Analysis of vertex inclusion probabilities in $\numlayers$-hop neighborhoods with node-wise sampling}
\label{sec:neighbor-access-analysis}

We analyze the following random process, which models neighborhood expansion in GNN architectures with node-wise sampling.
\begin{enumerate*}[label=(\roman*)]
  \item Start from a set of vertices rather than a single vertex.
  \item\label{itm:nhood-expand} For each vertex in the starting set, sample a set of direct neighbors without replacement, thereby expanding the walk frontier.
  \item Using the union of sampled neighborhoods as the new starting set, repeat step (ii). 
\end{enumerate*}
The process terminates after $\numlayers$ repetitions (i.e., hops).  
We assume that sampling is independent across hops and across vertices in the current-hop set.  That is, although each vertex samples among its direct neighbors without replacement, different vertices may sample the same direct neighbor.

We seek to estimate the vertex inclusion probabilities, or VIP values, in expanded neighborhoods that are obtained per the above random process.  The VIP values form a vector of probabilities $p(u)$ over all graph vertices.  We can express $p(u)$ via hop-wise VIP vectors $p\ofhop(u)$ that contain the probabilities that any vertex $u$ is sampled exactly $\hop$ hops away from the starting set.  (The hop-$\hop$ neighborhood contains the input vertices for the $\layer$-th GNN layer, $\layer = \numlayers - \hop$.)  A vertex $u$ is \emph{not} sampled at hop $\hop$ only if, for each of its neighbors $v \in \nhood_{1}(u)$, either $v$ was not sampled at hop $\hop - 1$ or $v$ was sampled but did not pick $u$ among its neighbors.  
The following \namecref{prop:prob-exp-nhood} shows how to propagate the initial-set probabilities through the graph to compute the VIP values.  We give a proof in \cref{app:proof-prob-exp-nhood-exact}.

\begin{proposition}
  [Vertex inclusion probabilities in minibatch neighborhood expansion with node-wise sampling]
  \label{prop:prob-exp-nhood}
  The probability that vertex $u \in \vertices$ appears in the node-wise sampled
  $\numlayers$-hop expanded neighborhood of any minibatch is
  \begin{gather}
    \label{eq:prob-exp-nhood-total}
    p(u) = 1 - \prod_{\hop=1}^{\numlayers} \! \big( 1 - p\ofhop(u) \big) ,
    \\
    \label{eq:prob-exp-nhood-layer}
    p\ofhop(u) = 1 - \prod_{v \in \nhood_{1}(u)} \! \big(
      1 - \weight\oflayer[\hop](u,v) \, p\ofhop[\hop-1](v)
    \big) ,
  \end{gather}
  where $p\ofhop[0](v)$ is the probability that $v$ is in the minibatch and
  $\weight\oflayer[\hop](u,v)$ is the transition probability that $v$ samples
  $u$ as a neighbor at hop $\hop = 1, \ldots, \numlayers$.
\end{proposition}

The VIP model of \cref{prop:prob-exp-nhood} applies to any initial sampling and hop-wise transition probability function for node-wise sampling. 
For example, if minibatches and vertex-wise neighbors are sampled uniformly at random without replacement as in GraphSAGE, then we have: $p\ofhop[0](u) = \batchsize / \setcard{\trainvertices}$ if $u \in \trainvertices$ and $0$ otherwise, where $\batchsize$ is the minibatch size and $\trainvertices$ is the set of training vertices; and $\weight\oflayer[\hop](u,v) = \min \setenum{1, \fanout\oflayer[\hop] / \deg(v)}$ if $u \in \nhood_{1}(v)$ and $0$ otherwise, where $\deg(v)$ is the (outgoing) degree of $v$.
Non-uniform neighbor sampling models are accommodated via the corresponding transition probability matrix or matrices.

The neighborhood expansion random process parametrizes a continuum between a random walk and full neighborhood expansion.  If the initial set size and layer-wise fanouts are equal to 1, it becomes a random walk.  If the fanouts are greater than the max in-degree of graph vertices, it becomes a full neighborhood expansion. 
The VIP model for the above two special cases is linear, while the generalized model in \cref{prop:prob-exp-nhood} is nonlinear; nonetheless, these models all have the same computational complexity: $O(\numlayers(\numedges + \numvertices))$.


\subsection{Vertex feature caching}
\label{sec:neighbor-access-communication-reduction}

We now show how to embody VIP analysis into a caching policy for minimizing communication among distributed machines and host-to-device data transfers in each machine.

\paragraph{Communication reduction}

We consider the optimal static caching policy, in the maximum-likelihood sense, for reducing inter-partition communication volume in distributed GNNs.  The policy is straightforward: each machine extends its local vertex feature storage with copies of the remote features that are most likely to be accessed by the machine, thereby minimizing the total expected communication volume.   
For some given initial partitioning --- and making no assumptions about the order in which minibatch and neighboring vertices are sampled --- this policy is directly related to the VIP analysis.  Specifically, we calculate partition-wise VIP values $p\ofpart(u)$ from the corresponding initial probabilities $p\ofhop[0]\ofpart(u)$.  The cache contents for the $\part$-th partition are then determined by ranking the remote vertices in order of decreasing $p\ofpart(u)$.

We measure the size of a remote-feature cache by its corresponding \defn{replication factor} $\repfactor$.  The replication factor is defined such that the number of cached feature vectors stored in each machine is $\repfactor \numvertices / \numparts$, where $\numparts$ is the number of partitions or machines.  That is, the fraction of cached to local vertices for each partition is $\repfactor$, and each feature vector is stored in $(1 + \repfactor)$ machines on average.  We may say that a partition or cache replicates a remote vertex to mean that it copies the corresponding vertex feature data.

\renewcommand{\textfraction}{0.05}

\begin{figure}
  \centering
  %
  %
  %
  \begin{subfigure}{\linewidth}
    \centering
    \includegraphics[width=\linewidth]{%
      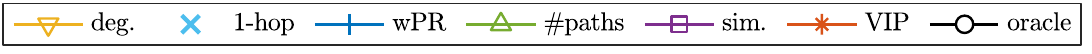}
  \end{subfigure}%
  \\[0.15em]
  %
  \begin{subfigure}{0.48\linewidth}
    \centering
    \includegraphics[width=\linewidth]{%
      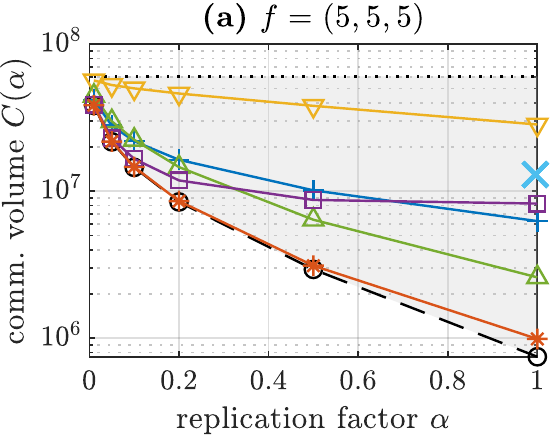}
  \end{subfigure}%
  \hspace*{\fill}
  %
  %
  \begin{subfigure}{0.48\linewidth}
    \centering
    \includegraphics[width=\linewidth]{%
      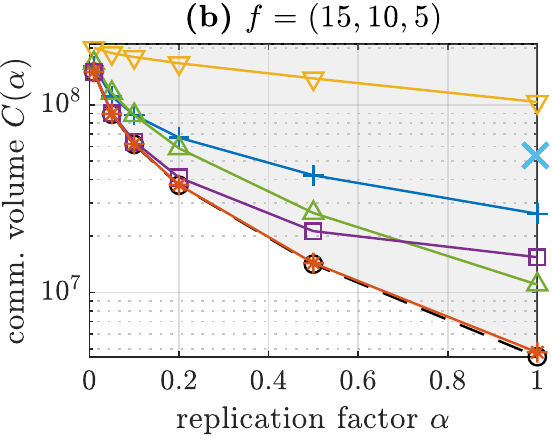}
  \end{subfigure}%
  \\[0.15em]
  %
  \begin{subfigure}{0.48\linewidth}
    \centering
    \includegraphics[width=\linewidth]{%
      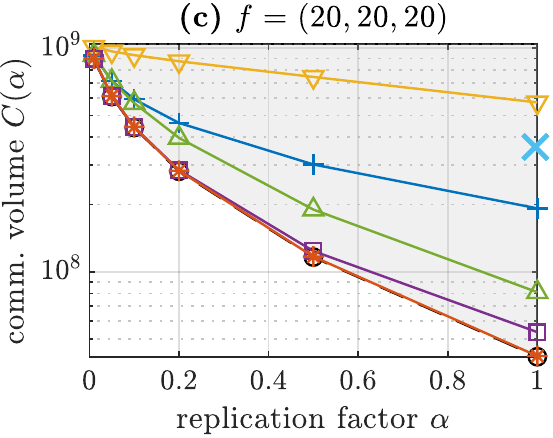}
  \end{subfigure}%
  \hspace*{\fill}
  %
  %
  \begin{subfigure}{0.48\linewidth}
    \centering
    \includegraphics[width=\linewidth]{%
      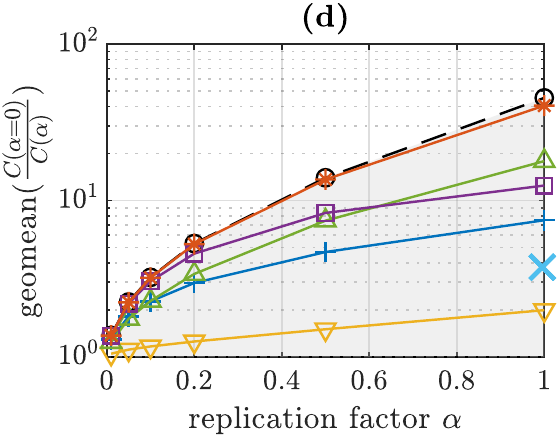}
  \end{subfigure}%
  %
  %
  \vspace{-2ex}
  \caption{Comparison of caching policies with respect to remote feature
    communication volume.
    %
    %
    GNN:~3-layer GraphSAGE, varying fanouts $\fanout$, minibatch size 1024. %
    Data set: ogbn-papers100M (undirected). %
    Partition: 8-way METIS with edge-cut objective plus edge and training/validation/test vertex balancing.
    \textbf{(a)}--\textbf{(c)}~Average per-epoch communication  in number of vertices over 100 epochs (log-scale).
    Lower is better.  The feasible region between the communication upper bound (no caching) and lower bound (oracle caching) is shaded gray.
    \textbf{(d)}~Geometric-mean improvement across fanouts, relative to no caching (log-scale).  Higher is better.%
  }
  \label{fig:cache-comm-vol-comparison-repfactor}
\end{figure}

\Cref{fig:cache-comm-vol-comparison-repfactor} compares a number of static feature-caching policies and provides evidence that the analytical VIP model of \cref{prop:prob-exp-nhood} effectively yields the optimal policy.
The specific policies being compared are as follows:
\begin{enumerate*}
  \item[``deg.''] ranks vertices by degree, after filtering out remote vertices that are not reachable from a partition's training set~\citep{PaGraph};
  \item[``1-hop''] replicates the entire $1$-hop halo of each partition;
  \item[``wPR''] ranks vertices by their score after 5 iterations of a weighted reverse-PageRank model~\cite{Min2021} with damping factor 0.85;
  \item[``\#paths''] ranks vertices by the number of paths with length $\le \numlayers$ that reach them from any local training vertex;
  \item[``sim.''] ranks vertices by empirical VIP estimates, measured by counting vertex-wise frequencies of access over 2 simulated training epochs~\cite{Yang2022}; 
  \item[``VIP''] ranks vertices by VIP values as per \cref{prop:prob-exp-nhood}; and
  \item[``oracle''] retroactively ranks vertices by their actual access frequencies after training, providing a lower bound on communication volume.%
  \footnote{With all caching policies, we calculate vertex rankings with respect to each partition.  This is different from the original setting in which some of these policies were introduced, where vertices are cached based on a single, global ranking score.}
\end{enumerate*}
We make the following observations.
\vspace{1em} 
\begin{itemize}[noitemsep,topsep=0pt,parsep=0.5\parskip,partopsep=0pt,leftmargin=*]
\item 
\emph{Optimality.}
The analytical VIP policy yields near-optimal communication volume (within 5\% of ``oracle'') across fanouts and replication factors.  The bigger gap ($\sim$30\%) for $\fanout = (5,5,5)$ and $\repfactor = 1.0$ is due to variance in the accesses of lower-ranked vertices; the gap narrows with more samples (e.g., bigger fanouts or more epochs).

\item
\emph{Communication reduction efficacy.}
VIP-driven caching is highly effective for reducing communication volume.  Compared to no caching, the VIP policy
achieves a geometric-mean reduction of \speedup{2.2}--\speedup{5.3} with small
replication factors $\repfactor \in [0.05, 0.20]$ and more than \speedup{10} with replication factors over $0.50$.  
Compared to the other caching
policies, it achieves consistently lower communication volume and its relative
improvement increases with the replication factor.

\item
\emph{Local information is not sufficient.}
The high-degree and 1-hop halo policies, which do not take neighborhood expansion into account, offer only marginal improvements over no caching.  Caching multi-hop neighbors will reduce communication further~\citep{PaGraph} but it will also blow up the replication factor.

\item \emph{Impact of tailoring the model to the expansion process.} 
The VIP-driven policy is up to \speedup{4} and \speedup{2} more effective than the ``wPR'' and ``\#paths'' policies, respectively.  Both of these policies take graph-structural expansion into account, but ``wPR'' is agnostic to the GNN fanout and number of layers, while ``\#paths'' does not directly account for sampling.

\item \emph{Benefits and limitations of empirical estimation.}
The empirical and analytical VIP estimates (``sim.'' and ``VIP'' policy, respectively) yield comparable results
for low replication factors and high fanouts, but the empirical
estimates become less effective as the replication factor increases or the
fanouts decrease.  For replication factors $0.50$ and $1.0$, the relative aggregate improvement of the analytical over the empirical policy is \speedup{1.6} and \speedup{3.2},
respectively.  The empirical approach has the benefit that it can be used for any sampling scheme, but it also requires increasingly
many samples to accurately estimate VIP values for infrequently accessed
vertices.

\end{itemize}

\paragraph{Host-to-device data transfer reduction}

VIP analysis can also be used to reduce the volume of host-to-device data
transfers for local features on each machine.  For example, assume that each
GPU can retain a fraction of the local feature data in memory
throughout the GNN computations (i.e., across all minibatches).  We may rank the local vertices in the $\part$-th partition by $p\ofpart(u)$ and keep the highest-ranking-vertex features on the GPU\@.  As with remote-vertex caching, this policy minimizes the expected data transfer volume due to local vertices in each partition.
   \clearpagedraft
\section{\oursystem: Fast and scalable distributed GNN training via caching and pipelining}
\seclabel{system-design}

This section describes the design of \oursystem, a fast and scalable system for performing distributed
minibatch training of GNNs on large partitioned datasets. 
The design of \oursystem is comprised of three majors components: 
\begin{description}[nosep]
	\item[Vertex reordering] A vertex ordering 
    based on the graph partitions and VIP values that facilitates efficient subpartitioning of local feature data between CPU and GPU to reduce host-to-device data transfers.
	\item[Data replication] An economical data replication strategy that uses VIP analysis to reduce communication volume for fetching remote features while using only a small amount of additional memory.
	\item[Pipelined communication] A pipelined distributed minibatch preparation system that hides latencies due to sampling, data transfers, and inter-machine communication of remote feature data.
\end{description}

These three techniques together
enable \oursystem to perform GNN
training on partitioned data-graphs while matching the efficiency of highly optimized GNN training systems that perform distributed computations with
full replication of vertex feature data across all machines. This enables
\oursystem to perform GNN training efficiently on large datasets where
full replication is impractical. 

As we shall discuss in our empirical evaluation in \secref{eval}, \oursystem achieves high performance for GNN training on large datasets even when operating in clusters with only modest hardware configurations. 

\subsection{Partitioning and reordering of vertex features}

Let us describe how \oursystem partitions the vertex features of a graph to reduce inter-machine
and host-to-device communication. \oursystem employs a two-level strategy for distributing the
vertex features across machines and devices, as illustrated in \figref{partitioned-frequency-ordered-data}.
The top level involves distributing vertex features based on a vertex partitioning of the graph,
and the bottom level involves dividing each partition's local vertex features between GPU and CPU memory.

\oursystem operates on graph data sets that are partitioned using a graph
partitioning algorithm such as METIS~\cite{metis}.  The graph partitions are
generated such that the training, validation, and test vertices are balanced. As a
heuristic to balance work-per-partition, an additional criterion is used
to balance the amount of edges per partition.
Each machine is assigned a subset of training/validation/test vertices and stores vertex feature data locally according to the machine's partition.

The graph is reordered such that: (a) indices for vertices in the same
partition are contiguous; and (b) vertices within a partition are ordered based on how beneficial it is for them to be stored on the GPU. Each machine stores a prefix of its ordered list of vertex features on the GPU, which when using VIP ordering corresponds to those vertices that are accessed most frequently by the local machine. This partitioning structure facilitates efficient determination of whether a vertex is remote or local, where
a local vertex is stored on the machine, and index calculations using
only a constant amount of additional memory. \figref{partitioned-frequency-ordered-data}
provides a graphical illustration of the VIP-reordered partitioned dataset.

\subsection{Replication of remote vertex features}

Each machine maintains a replica or static cache of features for vertices in remote partitions, where the size of the cache is determined by a given
replication factor $\repfactor$.  The vertices whose features are cached in each machine are determined by their VIP rank to reduce network communication, as described in \cref{sec:neighbor-access-communication-reduction}. 

Once the neighborhood sampling code prepares a batch, 
\oursystem partitions the vertices in the expanded neighborhood into local and remote vertices. Among the subset of remote vertices, an efficient lookup is performed using a hash table to determine whether the features for a remote vertex can be found in the local machine's cache. Only remote vertices that are not resident in the local cache proceed to be requested from remote machines.

\begin{figure}
	\includegraphics[trim={0.5cm 1.8cm 1.9cm 0.3cm},clip,width=\linewidth]{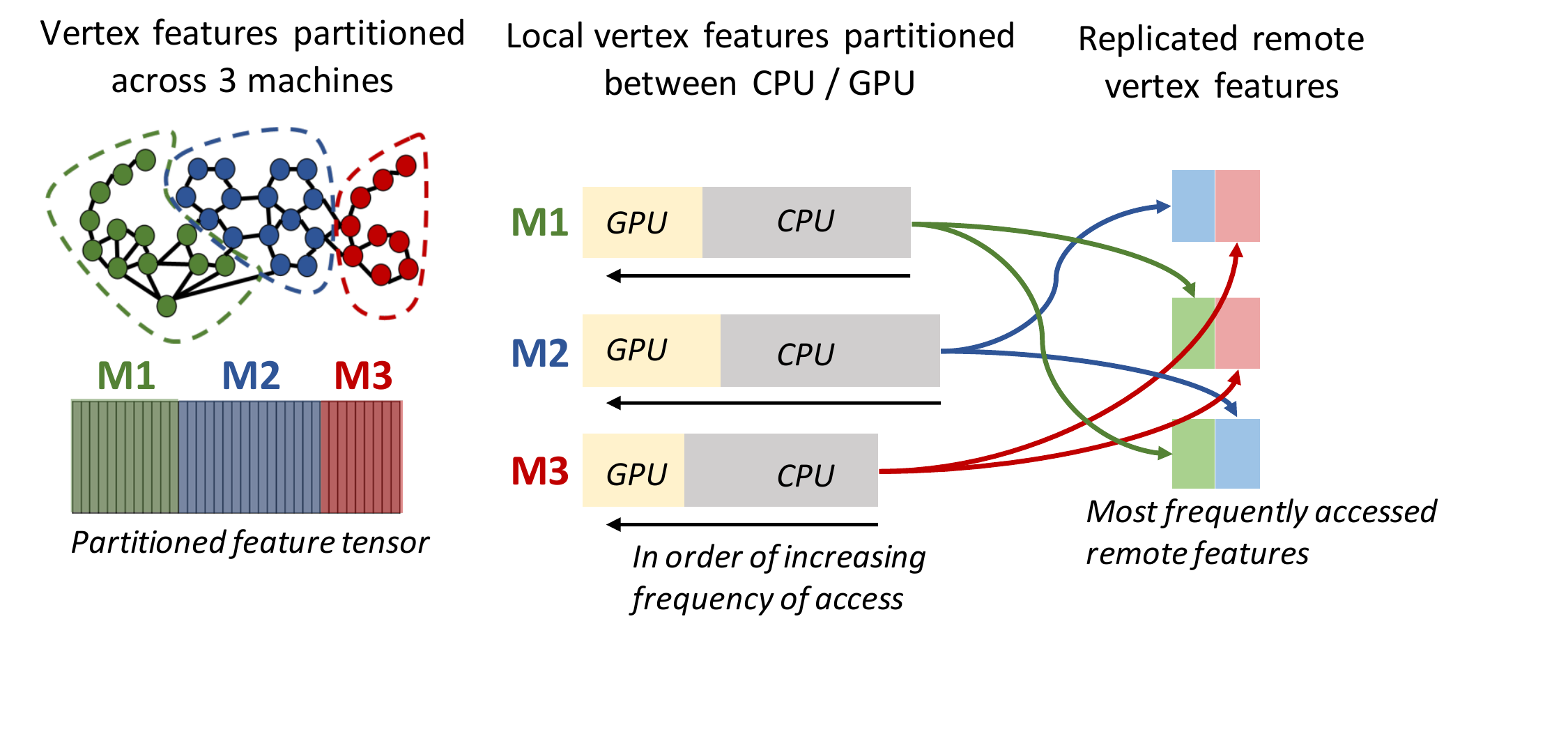}
    \vspace{-4ex}
	\caption{Illustration of the partitioned data set with VIP-driven local ordering and caching of remote features across machines.}
	\figlabel{partitioned-frequency-ordered-data}
    \vspace*{-2ex}
\end{figure}

\subsection{Minibatch preparation pipeline}

\oursystem implements a deep multi-stage pipeline 
which
enables overlap between neighborhood sampling,
host-to-device data transfers, network communication between machines, and
training computation. \oursystem's pipeline is more complex than the one used in
SALIENT~\citep{salient}, but it maintains the latter's usability and efficiency. \oursystem's
pipeline can be used in place of standard data loaders for GNNs by
changing just a few lines of code.

\textbf{Sampling and slicing} \oursystem uses a modified version of SALIENT's optimized neighborhood sampling and slicing code to prepare
minibatches using shared-memory parallelism. The original SALIENT batch-preparation code fused the neighborhood sampling and feature tensor slicing operations.
In \oursystem, these tensor-slicing
operations are only performed on the subset of the vertices in the sampled neighborhood that are stored locally in the
CPU memory of each machine. 
For the remaining vertices in the sampled neighborhood, \oursystem distinguishes between
vertices in the following categories: vertices belonging to the local partition whose features are stored on GPU; remote vertices
that are replicated on the local machine; and remote vertices that belong to other partitions.  

\textbf{Feature collection pipeline} We briefly summarize the pipeline operations. The expanded neighborhood of each minibatch may include remote vertices, remote vertices whose features were cached on the local machine following VIP analysis, and local vertices which are split across CPU and GPU storage. Immediately following sampling, vertices are categorized by their storage location so that vertex features are properly retrieved. Communication rounds interspersed with host-to-device and device-to-host transfers are enqueued with remote machines to coordinate retrieval of remote features. Replicated and local features partitioned across CPU and GPU storage are sliced locally. Finally, all features are concatenated into a single tensor and reordered for compatability with the message-flow graph generated during neighborhood sampling.

A total of $10$ mini-batches can be ``in-flight'' in the \oursystem pipeline at any time, each mini-batch being processed by a separate stage of the pipeline. Pipelining enables overlap between host-to-device data transfers and network communication,
and it also allows for hiding latencies related to CPU-GPU data transfers and inter-machine communication. A more detailed stage-by-stage description of \oursystem's pipeline is provided in \cref{sec:pipeline-stages}. 

         \clearpagedraft
\section{Evaluation}
\seclabel{eval}

In this section, we evaluate the performance of \oursystem and investigate the impact of its optimizations relating to
pipelining and use of VIP analysis (as developed in \secref{neighbor-access}). Additionally, we report end-to-end performance results,
model accuracies, and contextualize our performance in relation to existing distributed GNN training systems.

\begin{table}
  \vspace{-1ex}
  \caption{Summary of data sets. Edge counts reflect the number of edges in the graph after standard preprocessing (e.g., making the graph undirected). The \magcites data set is the papers-to-papers citation subgraph of the \textit{lsc-mag240} heterogeneous graph data set.}
  \label{tab:datasets}
  \vskip 0.05in
  \centering
  \small
  \begin{tabularx}{\linewidth}{@{} Xcccc @{}}
    \toprule
    \emph{Data Set} & \emph{\#Vertices} & \emph{\#Edges} & \emph{\#Feat.} & \emph{Train. / Val. / Test} \\
    \midrule
    products & 2.4M & 123M & 100 & 197K / 39K / 2.2M \\
    papers & 111M & 3.2B & 128 & 1.2M / 125K / 214K \\
    \magcitesfigs & 121M & 2.6B & 768 & 1.1M / 134K / 88K  \\
    \bottomrule
  \end{tabularx}
\vskip 0.05in
    \caption{GNN architecture hyperparameters for all experiments in \cref{sec:eval}. Fanout is for
    training. Batch size is per GPU.}
  \label{tab:GNNs}
  \vskip 0.05in
  \centering
   \small
  \setlength\tabcolsep{4pt}
  \begin{tabularx}{\linewidth}{@{} XXcccc @{}}
    \toprule
    \emph{Data Set} & \emph{GNN} & \emph{\#Layers} &\emph{Hidden Dim.} & \emph{Fanout} & \emph{Batch} \\
    \midrule
    products  & SAGE    & 3 &  256 & (15, 10, 5)  & 1024 \\
    papers    & SAGE    & 3 &  256 & (15, 10, 5)  & 1024 \\
    \magcitesfigs & SAGE     & 2 &  1024 & (25, 15)  & 1024 \\
    \bottomrule
  \end{tabularx}
\end{table}

\begin{figure}[t]
  \centering
	\includegraphics[width=\linewidth]{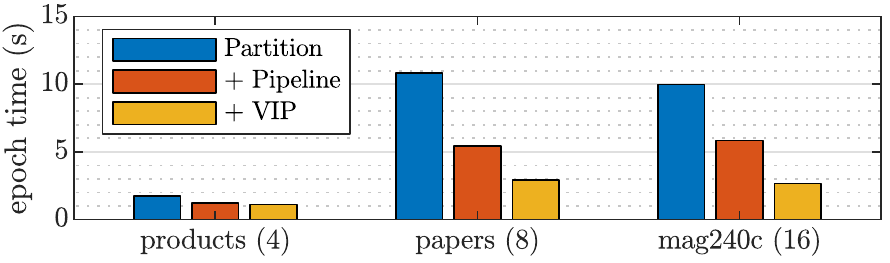}
    \vspace{-4ex}
	\caption{Impact of pipelining and VIP optimizations in
    \oursystem. Illustrates the performance improvements obtained by
    successively more optimized versions of \oursystem for \textit{products} (4
    partitions), \textit{papers} (8 partitions), and \magcites (16
    partitions). %
    The GNN architectures for each benchmark are listed in \cref{tab:GNNs}.
    The replication factors for VIP-based caching were $0.16$, $0.32$, and $0.32$ for \textit{products}, \textit{papers}, and \magcites, respectively.}
	\figlabel{ablation-study}
     \vspace{-2ex}
\end{figure}

\subsection{Experimental setup}
\seclabel{experimental-setup}
\textbf{Computing environment}
Our experiments were run on a cluster consisting of 16 
machines, each equipped with a 16-core (2-way hyperthreaded) AMD CPU with 128GB DRAM and 1 
NVIDIA A10G GPU with 24GB memory. The cluster was configured using the Amazon Web Services~(AWS) ParallelCluster software and employed fleets of AWS \textit{g5.8xlarge} instances. The network SLA specified for this instance type is 25Gbps. Distributed communications were implemented via PyTorch's DistributedDataParallel module with the NCCL backend. Since our cluster comprises machines with 1 GPU each, experiments with $K$ GPUs involve $K$ separate machines communicating over the network.

\textbf{Datasets}
We evaluate \oursystem\ on three standard benchmark data
sets: ogbn-products (\textit{products}), ogbn-papers100M 
 (\textit{papers})~\citep{Hu2020}, and
lsc-mag240 (\magcites)~\citep{hu2021ogb}. The graph and training set in these
data sets vary in size, with \textit{papers} and \magcites being two of the largest
open benchmarks at the time of this work. See Table~\ref{tab:datasets}
for detailed information. All graphs were made undirected (if originally not),
as is common practice. For \textit{mag240c}, we train on the homogeneous papers-to-papers component of the graph.

\textbf{GNN architectures}  
We evaluated \oursystem's performance using standard GraphSAGE~\citep{Hamilton2017} architectures
with each data set's most commonly used hyperparameters. \tabref{GNNs}
lists the GNN architectures and hyperparameters that were used for each data set.

\begin{figure}[t]
  \centering
        \begin{minipage}[l]{\columnwidth}
	\input{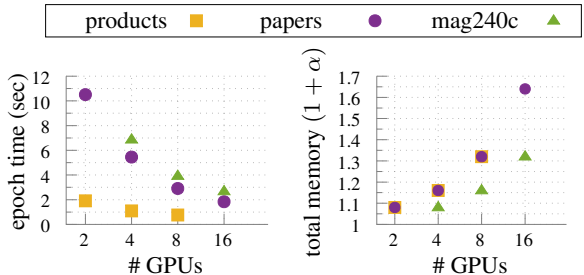}
	\end{minipage}
		\vspace{-4ex}
	\caption{Scalability and total memory used by all machines for \textit{products}, \textit{papers}, and \magcites. Total memory is plotted as a multiple of the unreplicated data set size $1+\alpha$ where $\alpha$ is the replication factor
 used for the $K$-GPU execution. }
	\figlabel{scalability}
    \vspace*{-1ex}
\end{figure}

\begin{figure}[t]
    \begin{minipage}[b]{0.35\columnwidth}
	\input{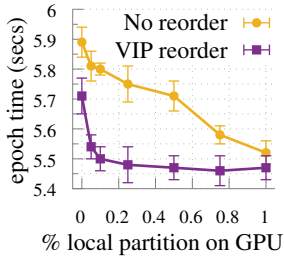}
    \end{minipage}%
    \hspace{\fill}%
    \begin{minipage}[b]{0.48\columnwidth}
	    \caption{Impact of VIP-based local vertex ordering on per-epoch runtime. Experiment run on $4$ GPUs with the \textit{papers} dataset and replication factor $\repfactor = 0.15$. Reported numbers are the mean over 10 epochs and vertical lines show the standard deviation.}
        \label{fig:cpu-vs-gpu-storage-impact}
    \end{minipage}
    \vspace*{-1ex}
\end{figure}

\subsection{Performance analysis of \oursystem}

We analyze the performance of \oursystem by investigating the impact of its component optimizations, and measuring end-to-end performance and scalability across three large graph data sets.

\textbf{Summary of performance improvements}
\figref{ablation-study} summarizes how \oursystem's optimizations result in increasingly better end-to-end training 
performance on \textit{products}, \textit{papers}, and \magcites in the distributed setting.
Significant improvements are observed for both the \textit{papers} and \magcites benchmarks. In relative terms, the \textit{papers} benchmark benefits equally from pipelining and VIP-based caching, while the \magcites benchmark benefits slightly more from caching on top of pipelining.  This is mainly due to the larger feature dimensionality for \magcites, which causes the communication of remote feature data to be relatively more throughput-bound than for \textit{papers}.


	%

\textbf{Scalability}
\figref{scalability} (left plot) illustrates the reduction in per-epoch runtime when using
\oursystem to execute the \textit{products}, \textit{papers}, and \magcites
benchmarks on $2$--$16$ GPUs.  For all three benchmarks, the reported runtimes
include warm-up time to fill up the pipeline at the start of
each training epoch, which results in diminished scalability once per-epoch
runtimes become less than a second. The \textit{products} benchmark achieves
\paperdata{\speedup{1.7}} speedup when going from $2$ to $4$ GPUs, but does not
significantly benefit from more scaling. The \textit{papers} benchmark achieves
\paperdata{\speedup{1.9}} speedup from $2$ to $4$ GPUs and another \speedup{1.9} from $4$ to $8$ GPUs. The \magcites
benchmark achieves \paperdata{\speedup{1.75}} speedup going from $4$ to $8$ GPUs, and \paperdata{\speedup{1.45}} speedup from
$8$ to $16$ GPUs. 
\oursystem's performance matches that of prior fast
systems which perform distributed GNN training with full replication~\citep{salient}, while
\oursystem uses substantially less memory. \figref{scalability} (right plot)
shows the total memory for storing vertex feature data (including replicated vertices) across all machines. Note that full
replication with $\numparts$ machines corresponds to replication factor
$\repfactor = \numparts-1$.

\paragraph{Local partition storage on CPU versus GPU}
\figref{cpu-vs-gpu-storage-impact} illustrates the performance impact of
increasing the percentage of the local partition stored on GPU while running the \textit{papers} benchmark. Given an
ordering of the vertices in the local partition and a percentage $\gpufactor\%$
of local data to store on GPU, \oursystem stores the first $\gpufactor\%$ of
vertex features on the GPU. The results labeled ``no reorder'' show an essentially linear reduction in
per-epoch runtime as a function of $\gpufactor\%$ when ranking local vertices
by their initial ordering.  The results labeled ``VIP reorder,'' on the other hand, show that when ranking local vertices by their VIP values, data transfer bottlenecks are effectively eliminated with as little as $10\%$ of the local partition data on the GPU.

\begin{figure}[t]
        \hspace{-3ex}\begin{minipage}[l]{0.49\columnwidth}
		\input{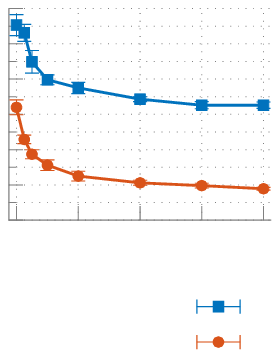}
	\end{minipage}
	\begin{minipage}[c]{0.49\columnwidth}
	\input{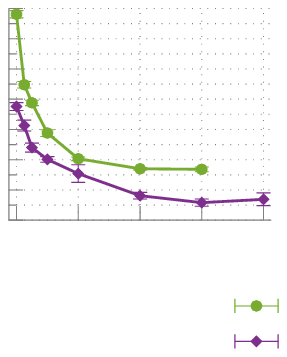}
	\end{minipage}\
	\vskip -0.15in
	\caption{Replication factor impact on per-epoch runtime. Results are shown for \textit{papers} (left) for $4$ and $8$ partitions, and for \magcites for $8$ and $16$ partitions for replication factors $\alpha$ varied between $0$ and $0.32$. The percentage of the local partition stored on GPU for this experiment was $90\%$ for \textit{papers} and $10\%$ for \magcites. Reported numbers are the mean value over 10 epochs and vertical bars indicate the standard deviation across epoch runtimes.} 
    \figlabel{replication-factor-impact}
    \vspace*{-1ex}
\end{figure}

\paragraph{Impact of replication factor}
\figref{replication-factor-impact} illustrates the impact on per-epoch runtime of increasing the number of replicated vertex features for \textit{papers} on 4 and 8 partitions (left) and on \magcites for 8 and 16 partitions (right). 
\figref{replication-factor-impact} shows that modest replication factors of $0.08$--$0.16$ and $0.16$--$0.32$ are sufficient for minimizing per-epoch runtime when using $4$ and $8$ partitions respectively on the \textit{papers} dataset.

\begin{figure}[t]
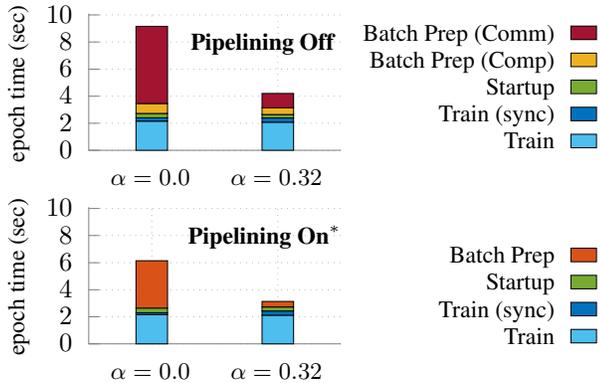

  \centering
        \begin{minipage}[l]{\columnwidth}
	\input{figs_epstex/performance_breakdown.tex}
	\end{minipage}
	\begin{minipage}[l]{\columnwidth}
	\input{figs_epstex/performance_breakdown_partialpipeline.tex}
	\end{minipage}
	\vspace{-4ex}
	\caption{Performance breakdown for \oursystem (with added synchronization) for an $8$-GPU execution of \textit{papers} with all local node features on GPU. The ``pipelining off'' breakdown is best able to isolate the time spent performing each operation. The ``pipelining on'' breakdown adds minimal synchronization to isolate distributed batch preparation from training computation. \textit{Batch Prep (Comm)} and \textit{Batch Prep (Comp)} are the time for communication and computation, respectively, in distributed batch preparation. These categories are combined into \textit{Batch Prep} for the ``pipelining on'' breakdown. \textit{Startup} is the time to prepare the first batch and/or fill the pipeline. \textit{Train (sync)} is the time for parallel workers to synchronize at the first collective operation of the training backward pass. \textit{Train} measures GPU computation for training.}
\figlabel{performance-breakdown}
\end{figure}

\paragraph{Performance breakdown for \oursystem}
\figref{performance-breakdown} illustrates the performance bottlenecks in \oursystem before and after incorporating pipelining and caching via VIP analysis.%
\footnote{\label{footnote:pipeline-on}The performance-breakdown experiments in \figref{performance-breakdown} store all local node features on the GPU.  Consequently, the ``pipelining off'' system with $\alpha=0$ is slightly faster in \figref{performance-breakdown} than in \tabref{performance-progression} where all local node features are on CPU.  The ``pipelining on'' system uses a version of \oursystem's pipeline with extra synchronization, resulting in modest slowdown, to facilitate the attribution of time among multiple network operations and concurrent computations running on different CUDA streams.}
A comparison of the ``pipelining on'' and ``pipelining off'' breakdowns for $\alpha=0$ illustrates that network communication is the primary bottleneck for distributed GNN training, and it remains the bottleneck even when communication is pipelined to overlap with computation. When using caching with $\alpha=0.32$, the time spent performing communication is sufficiently small so that it can be overlapped nearly perfectly with other computation in the program.%

\begin{figure}[t]
  \centering
        \begin{minipage}[l]{\columnwidth}
	\input{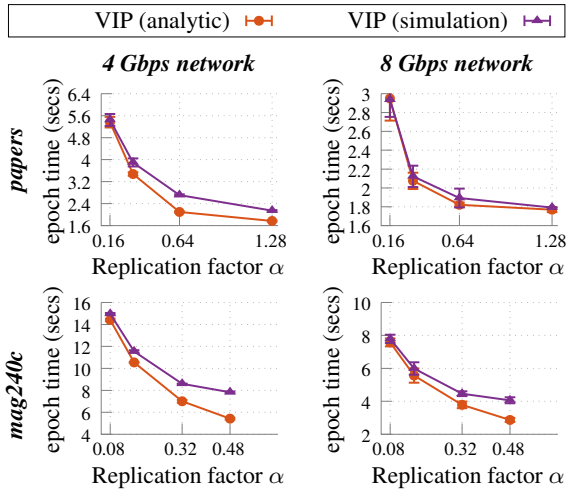}
	\end{minipage}
	\caption{Comparison of VIP-analytic versus VIP-simulation on end-to-end runtime on slow networks. Illustrates the per-epoch runtime of $16$-node executions on \textit{papers} and \magcites when using the VIP-analytic and VIP-simulation caching policies with different replication factors.}
	\figlabel{bw-experiment}
\end{figure}

\paragraph{Performance on a slow network}
\figref{bw-experiment} illustrates the performance of \oursystem on a slow network where higher replication factors are needed to alleviate communication bottlenecks. The slower network speeds were enforced using the Linux traffic control subsystem with the token-bucket filter (TBF) queuing discipline~\cite{hubert2002linux}. In this setting, we compare the two best
caching policies from \secref{neighbor-access}, VIP (analytic) and VIP (simulation). Recall from \secref{neighbor-access} that these two policies tend to perform very similarly for low replication factors, but diverge as the replication factor grows. On the \textit{papers} dataset, the relative gap between these two policies grows up to $30\%$ until $\alpha=0.64$ and then starts to shrink once communication ceases to bottleneck training time. For the \magcites dataset, which uses node features that are $6 \times$ larger than \textit{papers}, the relative gap between VIP (analytic) and VIP (simulation) is larger (up to $45\%$ with $\repfactor = 0.48$) and persistent. Whether or not VIP (analytic) is preferred to VIP (simulation) depends on the workload characteristics (e.g., fanout and size of node features) as well as the network bandwidth budget.

\subsection{Context for \oursystem's performance}

To conclude our empirical evaluation, we contextualize \oursystem's performance by reporting model accuracy results, discussing preprocessing overheads, and
comparing our performance to existing systems.

\paragraph{Model accuracy}
The optimizations in \oursystem do not impact model accuracy. \oursystem is
integrates with PyTorch Geometric~(PyG) and can use existing GNN models implemented
with PyG. Regardless, we report accuracies for \textit{products},
\textit{papers} and \magcites so that one can contextualize the performance
numbers. For computing accuracies, we executed
\oursystem on $8$ machines for $30$ epochs with a fixed learning rate of $0.001$ and a batch
size of $1024$ per machine.\footnote{These hyperparameters were not tuned for
accuracy, but are known to be reasonable values.} Sampling was used during inference with fanouts $(20,20,20)$ for \textit{papers} and \textit{products}, and $(25,15)$ for \magcites.
On \textit{products}, we
observed a test accuracy of \paperdata{$0.785$} for the $3$-layer SAGE architecture. On \textit{papers}, we observed a test accuracy of \paperdata{$0.646$} for the $3$-layer SAGE
architecture. On \magcites, the 2-layer GraphSAGE architecture achieves an
accuracy of \paperdata{$0.651$}.\footnote{Validation accuracy is reported instead of
``test-dev'' or ``test-challenge'' accuracy, which we did not measure at the
time of submission. Prior results published on the OGB leaderboard show that
this result is consistent for GraphSAGE on this dataset.}

\paragraph{Preprocessing overheads}
\oursystem's preprocessing overheads fall into two
categories: (a) runtime overheads incurred before each execution on a dataset; and, (b) 
dataset preparation overheads that are incurred only once and can be amortized over
multiple experiments. The runtime overheads for an $8$-node execution of \oursystem on the \textit{papers} dataset with replication factor $\alpha=0.32$ are as follows.  
Loading the dataset from disk takes approximately $10$ seconds. Computing
the VIP weights for fanouts $(15,10,5)$ takes $11.8$ seconds when implemented in PyTorch
with sparse transition weight matrices and dense hop-wise VIP-vectors. For large graphs, the VIP
computation code streams batches of matrix rows to the GPU, overlapping communication and data transfer. The communication of remote feature vectors and the related tensor slicing operations take about $22$ seconds. The dataset preprocessing overheads are highly dependent on the workflow used for partitioning graphs. Our (unoptimized) workflow for graph partitioning uses serial METIS to partition graphs and, additionally, uses machines with limited memory necessitating the use of swap files. In this setting, partitioning the \textit{papers} dataset takes $\sim\!2\;$hours and creating a reordered dataset from that partition takes $30$ minutes. Optimized workflows for partitioning can generate partitions substantially faster. For example, DistDGLv2~\cite{DistDGLv2} reports a $12$ minute runtime for partitioning when using ParMETIS~\cite{ParMETIS} and $4$ large multicores. \oursystem is agnostic to the source of the graph partitioning, and can be used in conjunction with more scalable graph partitioning codes.

\begin{table}[t]
  \caption{Comparison of \oursystem, DistDGL (public code), and
    DistDGLv2~\cite{DistDGLv2}. All numbers are reported for a $3$-layer GraphSAGE architecture
    with fanouts 15,10,5 and $256$ hidden nodes run on the \textit{papers} dataset. }
  \label{tab:distdgl-comparison}
  \vskip 0.05in
  \centering
  \small
  \begin{tabularx}{\linewidth}{@{} lcX @{}}
    \toprule
    \emph{System} & \emph{Time (s)}  & \emph{Notes} \\
    \midrule
    \oursystem & 2.9 & {\scriptsize 8 NVIDIA A10G GPUs, 25Gbps network throughput, \$19.5 / hr.} \\
    DistDGL & 37.0 & {\scriptsize Same hardware as is used by \oursystem. Example distributed code from Github\footnote{https://github.com/dmlc/dgl/}} \\
    DistDGLv2 & $\approx 5$ & {\scriptsize 64 NVIDIA T4 GPUs, 100Gbps network throughput, \$62.6 / hr.\footnotemark}\\
    \bottomrule
  \end{tabularx}
  \vskip -0.1in
\end{table}

\footnotetext{Hardware details obtained from DistDGLv2 paper~\cite{DistDGLv2}, and pricing information obtained from Amazon Web Services for the \emph{g4dn.metal} instance type reported as being used.}

\paragraph{Comparison to DistDGL}
We compared the performance of \oursystem to DistDGL on \textit{papers} 
using the GraphSAGE architecture. A summary of this comparison is provided in \tabref{distdgl-comparison}. The DistDGL code was obtained from DGL's public GitHub repository and executed in the same computing environment used to evaluate \oursystem. 
DistDGL's public code for distributed multi-GPU training was approximately \paperdata{12.7$\times$} slower than \oursystem on
$8$ GPUs (1 GPU per machine). Of course, DGL is under active development and its
support for efficient GNN training (especially in distributed environments) is evolving
rapidly. As such, we also report performance results from the recent work ``DistDGLv2''~\cite{DistDGLv2} 
that describes innovations in the DistDGL system for improving distributed GNN
training performance. Although not all of these innovations are publicly available, their reported performance can be used to
contextualize the performance we achieve with \oursystem.  A full comparison of the hardware used by DistDGLv2 and \oursystem shows that \oursystem consistently achieves similar (often better) performance than what was reported for DistDGLv2, while using substantially less resources. \oursystem's $8$-GPU runtime is approximately
\paperdata{$1.7 \times$} faster than DistDGLv2's reported numbers for executing a training epoch on \textit{papers} using $64$ GPUs. This faster per-epoch time is achieved with $8\times$ fewer GPUs, $4\times$ smaller network bandwidth, and $3.2\times$ cheaper hardware.
            \clearpagedraft
\section{Conclusions and future work}

We have presented a distributed multi-GPU system, \oursystem, for GNN training and inference on massive graphs. This system is built on top of SALIENT, a prior state-of-the-art system that attains efficiency and scalability through fast sampling and pipelining, at the cost of full data replication on all machines. In \oursystem, we distribute the vertex feature data and address the resulting communication bottleneck through analyzing the access pattern of the out-of-machine (i.e., remote) vertices and proposing a caching strategy that replicates a small amount of the most frequently accessed vertex features. Together with a deep pipelining of all operations from communication to computation, \oursystem retains the efficiency and scalability of SALIENT while consuming only a fraction of the storage required by SALIENT.

An avenue of future work is to further apply the access pattern analysis to
improve the initial graph partitioning, with an aim of reducing the
communication volume orthogonally to the use of caching. This would require
incorporating the vertex inclusion probabilities in the graph partitioning objective, on top of edge cuts and load balancing. Additionally, a hierarchical graph partitioning may better leverage the higher intra-machine bandwidth among GPUs than inter-machine communication. 
Another line of future work is to explore distributing the graph structure across machines. Distributing the graph incurs nontrivial challenges in the multi-round communication of node features, but resolving this challenge will further reduce memory consumption and bears the potential of handling graphs that are orders of magnitudes larger than the current largest benchmarks.

\section*{Acknowledgements}


This work was conducted on the SuperCloud computing cluster
\url{https://supercloud.mit.edu} and the Satori computing cluster
\url{https://mit-satori.github.io}. This research was sponsored by
MIT-IBM Watson AI Lab and in part by the United States Air Force
Research Laboratory and the United States Air Force Artificial
Intelligence Accelerator and was accomplished under Cooperative
Agreement Number FA8750-19-2-1000.  The views and conclusions
contained in this document are those of the authors and should not be
interpreted as representing the official policies, either expressed or
implied, of the United States Air Force or the U.S. Government.  The
U.S. Government is authorized to reproduce and distribute reprints for
Government purposes notwithstanding any copyright notation herein.

\bibliography{reference}

\begin{thebibliography}{44}
\providecommand{\natexlab}[1]{#1}
\providecommand{\url}[1]{\texttt{#1}}
\expandafter\ifx\csname urlstyle\endcsname\relax
  \providecommand{\doi}[1]{doi: #1}\else
  \providecommand{\doi}{doi: \begingroup \urlstyle{rm}\Url}\fi

\bibitem[Abadi et~al.(2015)Abadi, Agarwal, Barham, Brevdo, Chen, Citro,
  Corrado, Davis, Dean, Devin, Ghemawat, Goodfellow, Harp, Irving, Isard, Jia,
  Jozefowicz, Kaiser, Kudlur, Levenberg, Man\'{e}, Monga, Moore, Murray, Olah,
  Schuster, Shlens, Steiner, Sutskever, Talwar, Tucker, Vanhoucke, Vasudevan,
  Vi\'{e}gas, Vinyals, Warden, Wattenberg, Wicke, Yu, and Zheng]{Abadi2015}
Abadi, M., Agarwal, A., Barham, P., Brevdo, E., Chen, Z., Citro, C., Corrado,
  G.~S., Davis, A., Dean, J., Devin, M., Ghemawat, S., Goodfellow, I., Harp,
  A., Irving, G., Isard, M., Jia, Y., Jozefowicz, R., Kaiser, L., Kudlur, M.,
  Levenberg, J., Man\'{e}, D., Monga, R., Moore, S., Murray, D., Olah, C.,
  Schuster, M., Shlens, J., Steiner, B., Sutskever, I., Talwar, K., Tucker, P.,
  Vanhoucke, V., Vasudevan, V., Vi\'{e}gas, F., Vinyals, O., Warden, P.,
  Wattenberg, M., Wicke, M., Yu, Y., and Zheng, X.
\newblock {TensorFlow}: Large-scale machine learning on heterogeneous systems,
  2015.
\newblock https://www.tensorflow.org/.

\bibitem[Bottou et~al.(2018)Bottou, Curtis, and Nocedal]{Bottou2018}
Bottou, L., Curtis, F.~E., and Nocedal, J.
\newblock Optimization methods for large-scale machine learning.
\newblock \emph{SIAM Rev.}, 60\penalty0 (2):\penalty0 223--311, 2018.

\bibitem[Cai et~al.(2021)Cai, Yan, Wu, Ma, Cheng, and Yu]{DGCL}
Cai, Z., Yan, X., Wu, Y., Ma, K., Cheng, J., and Yu, F.
\newblock Dgcl: An efficient communication library for distributed gnn
  training.
\newblock In \emph{Proceedings of the Sixteenth European Conference on Computer
  Systems}, EuroSys '21, pp.\  130–144, New York, NY, USA, 2021. Association
  for Computing Machinery.
\newblock ISBN 9781450383349.
\newblock \doi{10.1145/3447786.3456233}.
\newblock URL \url{https://doi.org/10.1145/3447786.3456233}.

\bibitem[Chen \& Luss(2018)Chen and Luss]{Chen2018a}
Chen, J. and Luss, R.
\newblock Stochastic gradient descent with biased but consistent gradient
  estimators.
\newblock Preprint arXiv:1807.11880, 2018.

\bibitem[Chen et~al.(2018)Chen, Ma, and Xiao]{Chen2018}
Chen, J., Ma, T., and Xiao, C.
\newblock {FastGCN}: Fast learning with graph convolutional networks via
  importance sampling.
\newblock In \emph{ICLR}, 2018.

\bibitem[Chiang et~al.(2019)Chiang, Liu, Si, Li, Bengio, and Hsieh]{Chiang2019}
Chiang, W.-L., Liu, X., Si, S., Li, Y., Bengio, S., and Hsieh, C.-J.
\newblock Cluster-{GCN}: An efficient algorithm for training deep and large
  graph convolutional networks.
\newblock In \emph{KDD}, 2019.

\bibitem[Dong et~al.(2021)Dong, Zheng, Yang, and Karypis]{Dong2021}
Dong, J., Zheng, D., Yang, L.~F., and Karypis, G.
\newblock Global neighbor sampling for mixed {CPU-GPU} training on giant
  graphs.
\newblock In \emph{KDD}, 2021.

\bibitem[Gandhi \& Iyer(2021)Gandhi and Iyer]{Gandhi2021}
Gandhi, S. and Iyer, A.~P.
\newblock {P3}: Distributed deep graph learning at scale.
\newblock In \emph{OSDI}, 2021.

\bibitem[Gilmer et~al.(2017)Gilmer, Schoenholz, Riley, Vinyals, and
  Dahl]{Gilmer2017}
Gilmer, J., Schoenholz, S.~S., Riley, P.~F., Vinyals, O., and Dahl, G.~E.
\newblock Neural message passing for quantum chemistry.
\newblock In \emph{ICML}, 2017.

\bibitem[Hamilton et~al.(2017)Hamilton, Ying, and Leskovec]{Hamilton2017}
Hamilton, W.~L., Ying, R., and Leskovec, J.
\newblock Inductive representation learning on large graphs.
\newblock In \emph{NIPS}, 2017.

\bibitem[Hoang et~al.(2021)Hoang, Chen, Lee, Dathathri, Gill, and
  Pingali]{Hoang2021}
Hoang, L., Chen, X., Lee, H., Dathathri, R., Gill, G., and Pingali, K.
\newblock Efficient distribution for deep learning on large graphs,.
\newblock In \emph{GNNSys}, 2021.

\bibitem[Hu et~al.(2020)Hu, Fey, Zitnik, Dong, Ren, Liu, Catasta, and
  Leskovec]{Hu2020}
Hu, W., Fey, M., Zitnik, M., Dong, Y., Ren, H., Liu, B., Catasta, M., and
  Leskovec, J.
\newblock Open graph benchmark: Datasets for machine learning on graphs.
\newblock Preprint arXiv:2005.00687, 2020.

\bibitem[Hu et~al.(2021)Hu, Fey, Ren, Nakata, Dong, and Leskovec]{hu2021ogb}
Hu, W., Fey, M., Ren, H., Nakata, M., Dong, Y., and Leskovec, J.
\newblock Ogb-lsc: A large-scale challenge for machine learning on graphs.
\newblock \emph{arXiv preprint arXiv:2103.09430}, 2021.

\bibitem[Hubert et~al.(2002)]{hubert2002linux}
Hubert, B. et~al.
\newblock Linux advanced routing \& traffic control howto.
\newblock \emph{Netherlabs BV}, 1:\penalty0 99--107, 2002.

\bibitem[Jia et~al.(2020)Jia, Lin, Gao, Zaharia, and Aiken]{Jia2020}
Jia, Z., Lin, S., Gao, M., Zaharia, M., and Aiken, A.
\newblock Improving the accuracy, scalability, and performance of graph neural
  networks with {Roc}.
\newblock In \emph{MLSys}, 2020.

\bibitem[Kaler et~al.(2022)Kaler, Stathas, Ouyang, Iliopoulos, Schardl,
  Leiserson, and Chen]{salient}
Kaler, T., Stathas, N., Ouyang, A., Iliopoulos, A.-S., Schardl, T., Leiserson,
  C.~E., and Chen, J.
\newblock Accelerating training and inference of graph neural networks with
  fast sampling and pipelining.
\newblock \emph{Proceedings of Machine Learning and Systems}, 4:\penalty0
  172--189, 2022.

\bibitem[Karypis \& Kumar(1997)Karypis and Kumar]{metis}
Karypis, G. and Kumar, V.
\newblock Metis: A software package for partitioning unstructured graphs,
  partitioning meshes, and computing fill-reducing orderings of sparse
  matrices.
\newblock 1997.

\bibitem[Karypis et~al.(1998)Karypis, Schloegel, and Kumar]{ParMETIS}
Karypis, G., Schloegel, K., and Kumar, V.
\newblock Parmetis---parallel graph partitioning and sparse matrix ordering
  library, version 2.0.
\newblock \emph{Univ. of Minnesota, Minneapolis, MN}, 10, 1998.

\bibitem[Kipf \& Welling(2017)Kipf and Welling]{Kipf2017}
Kipf, T.~N. and Welling, M.
\newblock Semi-supervised classification with graph convolutional networks.
\newblock In \emph{ICLR}, 2017.

\bibitem[Li et~al.(2016)Li, Tarlow, Brockschmidt, and Zemel]{Li2016}
Li, Y., Tarlow, D., Brockschmidt, M., and Zemel, R.
\newblock Gated graph sequence neural networks.
\newblock In \emph{ICLR}, 2016.

\bibitem[Li et~al.(2018)Li, Yu, Shahabi, and Liu]{Li2018}
Li, Y., Yu, R., Shahabi, C., and Liu, Y.
\newblock Diffusion convolutional recurrent neural network: Data-driven traffic
  forecasting.
\newblock In \emph{ICLR}, 2018.

\bibitem[Lin et~al.(2020)Lin, Li, Miao, Liu, and Xu]{PaGraph}
Lin, Z., Li, C., Miao, Y., Liu, Y., and Xu, Y.
\newblock Pagraph: Scaling gnn training on large graphs via computation-aware
  caching.
\newblock In \emph{Proceedings of the 11th ACM Symposium on Cloud Computing},
  SoCC '20, pp.\  401–415, New York, NY, USA, 2020. Association for Computing
  Machinery.
\newblock ISBN 9781450381376.
\newblock \doi{10.1145/3419111.3421281}.
\newblock URL \url{https://doi.org/10.1145/3419111.3421281}.

\bibitem[Liu et~al.(2020)Liu, Wu, Zhang, Zhou, Yang, Song, and
  Qi]{liuBanditSamplers2020}
Liu, Z., Wu, Z., Zhang, Z., Zhou, J., Yang, S., Song, L., and Qi, Y.
\newblock Bandit samplers for training graph neural networks.
\newblock In \emph{Advances in {{Neural Information Processing Systems}}},
  volume~33, pp.\  6878--6888, 2020.

\bibitem[Ma et~al.(2019)Ma, Yang, Miao, Xue, Wu, Zhou, and Dai]{Ma2019}
Ma, L., Yang, Z., Miao, Y., Xue, J., Wu, M., Zhou, L., and Dai, Y.
\newblock {NeuGraph}: Parallel deep neural network computation on large graphs.
\newblock In \emph{USENIX ATC}, 2019.

\bibitem[Md et~al.(2021)Md, Misra, Ma, Mohanty, Georganas, Heinecke, Kalamkar,
  Ahmed, and Avancha]{md2021distgnn}
Md, V., Misra, S., Ma, G., Mohanty, R., Georganas, E., Heinecke, A., Kalamkar,
  D., Ahmed, N.~K., and Avancha, S.
\newblock Distgnn: Scalable distributed training for large-scale graph neural
  networks.
\newblock In \emph{Proceedings of the International Conference for High
  Performance Computing, Networking, Storage and Analysis}, pp.\  1--14, 2021.

\bibitem[Min et~al.(2021)Min, Wu, Huang, Hidayetoğlu, Xiong, Ebrahimi, Chen,
  and mei Hwu]{Min2021}
Min, S.~W., Wu, K., Huang, S., Hidayetoğlu, M., Xiong, J., Ebrahimi, E., Chen,
  D., and mei Hwu, W.
\newblock Large graph convolutional network training with {GPU}-oriented data
  communication architecture.
\newblock Preprint arXiv:2103.03330, 2021.

\bibitem[Mohoney et~al.(2021)Mohoney, Waleffe, Xu, Rekatsinas, and
  Venkataraman]{marius}
Mohoney, J., Waleffe, R., Xu, H., Rekatsinas, T., and Venkataraman, S.
\newblock Marius: Learning massive graph embeddings on a single machine.
\newblock In \emph{15th {USENIX} Symposium on Operating Systems Design and
  Implementation ({OSDI} 21)}, pp.\  533--549. {USENIX} Association, July 2021.
\newblock ISBN 978-1-939133-22-9.
\newblock URL
  \url{https://www.usenix.org/conference/osdi21/presentation/mohoney}.

\bibitem[Paszke et~al.(2019)Paszke, Gross, Massa, Lerer, Bradbury, Chanan,
  Killeen, Lin, Gimelshein, Antiga, Desmaison, Kopf, Yang, DeVito, Raison,
  Tejani, Chilamkurthy, Steiner, Fang, Bai, and Chintala]{Paszke2019}
Paszke, A., Gross, S., Massa, F., Lerer, A., Bradbury, J., Chanan, G., Killeen,
  T., Lin, Z., Gimelshein, N., Antiga, L., Desmaison, A., Kopf, A., Yang, E.,
  DeVito, Z., Raison, M., Tejani, A., Chilamkurthy, S., Steiner, B., Fang, L.,
  Bai, J., and Chintala, S.
\newblock {PyTorch}: An imperative style, high-performance deep learning
  library.
\newblock In \emph{NIPS}, 2019.

\bibitem[Ramezani et~al.(2020)Ramezani, Cong, Mahdavi, Sivasubramaniam, and
  Kandemir]{Ramezani2020}
Ramezani, M., Cong, W., Mahdavi, M., Sivasubramaniam, A., and Kandemir, M.
\newblock {GCN} meets {GPU}: Decoupling ``when to sample'' from ``how to
  sample''.
\newblock In \emph{NeurIPS}, 2020.

\bibitem[Veli\v{c}kovi\'{c} et~al.(2018)Veli\v{c}kovi\'{c}, Cucurull, Casanova,
  Romero, Li\`{o}, and Bengio]{Velickovic2018}
Veli\v{c}kovi\'{c}, P., Cucurull, G., Casanova, A., Romero, A., Li\`{o}, P.,
  and Bengio, Y.
\newblock Graph attention networks.
\newblock In \emph{ICLR}, 2018.

\bibitem[Waleffe et~al.(2022)Waleffe, Mohoney, Rekatsinas, and
  Venkataraman]{Marius++}
Waleffe, R., Mohoney, J., Rekatsinas, T., and Venkataraman, S.
\newblock Marius++: Large-scale training of graph neural networks on a single
  machine.
\newblock \emph{arXiv preprint arXiv:2202.02365}, 2022.

\bibitem[Wan et~al.(2022)Wan, Li, Li, Kim, and Lin]{BNSGCN}
Wan, C., Li, Y., Li, A., Kim, N.~S., and Lin, Y.
\newblock Bns-gcn: Efficient full-graph training of graph convolutional
  networks with partition-parallelism and random boundary node sampling.
\newblock In Marculescu, D., Chi, Y., and Wu, C. (eds.), \emph{Proceedings of
  Machine Learning and Systems}, volume~4, pp.\  673--693, 2022.
\newblock URL
  \url{https://proceedings.mlsys.org/paper/2022/file/d1fe173d08e959397adf34b1d77e88d7-Paper.pdf}.

\bibitem[Wang et~al.(2021{\natexlab{a}})Wang, Yin, Tian, Yang, Chen, Yu, Yao,
  and Zhou]{Wang2021}
Wang, L., Yin, Q., Tian, C., Yang, J., Chen, R., Yu, W., Yao, Z., and Zhou, J.
\newblock {FlexGraph}: a flexible and efficient distributed framework for {GNN}
  training.
\newblock In \emph{EuroSys}, 2021{\natexlab{a}}.

\bibitem[Wang et~al.(2021{\natexlab{b}})Wang, Feng, Li, Li, Deng, Xie, and
  Ding]{Wang2021a}
Wang, Y., Feng, B., Li, G., Li, S., Deng, L., Xie, Y., and Ding, Y.
\newblock {GNNAdvisor}: An adaptive and efficient runtime system for {GNN}
  acceleration on {GPUs}.
\newblock In \emph{OSDI}, 2021{\natexlab{b}}.

\bibitem[Weber et~al.(2019)Weber, Domeniconi, Chen, Weidele, Bellei, Robinson,
  and Leiserson]{Weber2019}
Weber, M., Domeniconi, G., Chen, J., Weidele, D. K.~I., Bellei, C., Robinson,
  T., and Leiserson, C.~E.
\newblock Anti-money laundering in {B}itcoin: Experimenting with graph
  convolutional networks for financial forensics.
\newblock In \emph{KDD Workshop on Anomaly Detection in Finance}, 2019.

\bibitem[Wu et~al.(2021)Wu, Ma, Cai, Jin, Li, Zheng, Cheng, and Yu]{Wu2021}
Wu, Y., Ma, K., Cai, Z., Jin, T., Li, B., Zheng, C., Cheng, J., and Yu, F.
\newblock {Seastar}: vertex-centric programming for graph neural networks.
\newblock In \emph{EuroSys}, 2021.

\bibitem[Xu et~al.(2019)Xu, Hu, Leskovec, and Jegelka]{Xu2019}
Xu, K., Hu, W., Leskovec, J., and Jegelka, S.
\newblock How powerful are graph neural networks?
\newblock In \emph{ICLR}, 2019.

\bibitem[Yang et~al.(2022)Yang, Tang, Song, Wang, Yin, Chen, Yu, and
  Zhou]{Yang2022}
Yang, J., Tang, D., Song, X., Wang, L., Yin, Q., Chen, R., Yu, W., and Zhou, J.
\newblock {{GNNLab}}: {{A}} factored system for sample-based {{GNN}} training
  over {{GPUs}}.
\newblock In \emph{EuroSys}, pp.\  417--434, 2022.

\bibitem[Ying et~al.(2018)Ying, He, Chen, Eksombatchai, Hamilton, and
  Leskovec]{Ying2018}
Ying, R., He, R., Chen, K., Eksombatchai, P., Hamilton, W.~L., and Leskovec, J.
\newblock Graph convolutional neural networks for web-scale recommender
  systems.
\newblock In \emph{KDD}, 2018.

\bibitem[Zeng et~al.(2020)Zeng, Zhou, Srivastava, Kannan, and
  Prasanna]{Zeng2020}
Zeng, H., Zhou, H., Srivastava, A., Kannan, R., and Prasanna, V.
\newblock {GraphSAINT}: Graph sampling based inductive learning method.
\newblock In \emph{ICLR}, 2020.

\bibitem[Zheng et~al.(2020)Zheng, Ma, Wang, Zhou, Su, Song, Gan, Zhang, and
  Karypis]{Zheng2020}
Zheng, D., Ma, C., Wang, M., Zhou, J., Su, Q., Song, X., Gan, Q., Zhang, Z.,
  and Karypis, G.
\newblock {{DistDGL}}: Distributed graph neural network training for
  billion-scale graphs.
\newblock In \emph{IA3}, 2020.

\bibitem[Zheng et~al.(2021)Zheng, Song, Yang, LaSalle, and Karypis]{Zheng2021}
Zheng, D., Song, X., Yang, C., LaSalle, D., and Karypis, G.
\newblock Distributed hybrid {CPU} and {GPU} training for graph neural networks
  on billion-scale graphs.
\newblock Preprint arXiv:2112.15345, 2021.

\bibitem[Zheng et~al.(2022)Zheng, Song, Yang, LaSalle, and Karypis]{DistDGLv2}
Zheng, D., Song, X., Yang, C., LaSalle, D., and Karypis, G.
\newblock Distributed hybrid cpu and gpu training for graph neural networks on
  billion-scale heterogeneous graphs.
\newblock In \emph{Proceedings of the 28th ACM SIGKDD Conference on Knowledge
  Discovery and Data Mining}, KDD '22, pp.\  4582–4591, New York, NY, USA,
  2022. Association for Computing Machinery.
\newblock ISBN 9781450393850.
\newblock \doi{10.1145/3534678.3539177}.
\newblock URL \url{https://doi.org/10.1145/3534678.3539177}.

\bibitem[Zou et~al.(2019)Zou, Hu, Wang, Jiang, Sun, and Gu]{Zou2019}
Zou, D., Hu, Z., Wang, Y., Jiang, S., Sun, Y., and Gu, Q.
\newblock Layer-dependent importance sampling for training deep and large graph
  convolutional networks.
\newblock In \emph{NeurIPS}, 2019.

\end{thebibliography}
\bibliographystyle{mlsys2023}


\clearpage
\appendix

\section{Artifact Appendix}

\subsection{Abstract}
 
This section describes the software artifacts associated with this paper. The code is distributed via GitHub at \url{https://github.com/MITIBMxGraph/SALIENT_plusplus_artifact} and can be
used to perform the experiments presented in the paper. 
We provide scripts to run: (a) the simulation experiments to compare different caching policies and generate data for \figref{cache-comm-vol-comparison-repfactor}; and (b) the distributed
experiments that produce data for \cref{tab:performance-progression,fig:ablation-study,fig:scalability,fig:cpu-vs-gpu-storage-impact,fig:replication-factor-impact}.
Detailed instructions for running these scripts are provided in a dedicated README file for artifact evaluation located at \url{https://github.com/MITIBMxGraph/SALIENT_plusplus_artifact/blob/main/README.md}.

\subsection{Artifact check-list (meta-information)}

\begin{itemize}[leftmargin=*]
  \item \textbf{Algorithm:} \oursystem distributed training algorithms for GNNs and VIP analysis algorithms for caching policies for GNNs.
  \item \textbf{Program:} PyTorch, CUDA
  \item \textbf{Compilation:} \texttt{gcc/g++} version 7 or greater; \texttt{nvcc} version 11.
  \item \textbf{Data set:} Node classification benchmark data sets from OGB.
  \item \textbf{Run-time environment:} Ubuntu 20.04 (or modern Linux distribution) with NVIDIA drivers installed.
  \item \textbf{Hardware:} NVIDIA GPU with sufficient memory. Distributed experiments require SLURM cluster with GPU nodes. Simulation experiments require up to 200 GB of main memory.
  \item \textbf{Experiments:} Local simulation experiments for comparing different caching policies in terms of remote vertex communication volume, and distributed experiments for analyzing the impact on per-epoch training time of different optimizations, varying replication factor, and varying percentage of data stored in-memory on GPU. 
  \item \textbf{How much disk space required (approximately)?:} 1.5~TB for all experiments, 100~GB for a substantial subset of experiments.
  \item \textbf{How much time is needed to prepare workflow (approximately)?:} 1--2 hours with prior experience and access to hardware/clusters.
  \item \textbf{How much time is needed to complete experiments (approximately)?:}
  4--12 hours for all experiments if using pre-processed datasets.
  \item \textbf{Publicly available?:} Yes
  \item \textbf{Code licenses:} Apache License 2.0
  \item \textbf{Data licenses:} Amazon license and ODC-BY.
  \item \textbf{Archive DOI:} \url{https://doi.org/10.5281/zenodo.7889203}
\end{itemize}

\subsection{Description}

\subsubsection{How delivered}

The code is available on GitHub at \url{https://github.com/MITIBMxGraph/SALIENT_plusplus_artifact}. Within the repository, scripts for streamlining the process of exercising the artifact are provided in the \verb|experiments| directory.  Instructions for running the scripts can be found in the repo's top-level README document at \url{https://github.com/MITIBMxGraph/SALIENT_plusplus_artifact/blob/main/README.md}.

\subsubsection{Hardware dependencies}

The minimum requirements for exercising the software artifact are as follows.

The simulation experiments can be performed on the ogbn-papers100M dataset using a single machine (with or without a GPU) that has 160GB or more of main memory.  On machines with lower memory capacity, it is possible to run these experiments by using swapfiles, ideally on fast SSDs.  Simulation experiments on the smaller ogbn-products dataset require less than 10 GB of memory.

The distributed multi-GPU experiments require access to either a SLURM cluster with GPU nodes or a single machine with multiple GPUs. Those opting to use a single machine with multiple GPUs should pass the appropriate flags to experimental scripts to indicate they are running scripts locally. Access to a SLURM cluster with GPU nodes may be obtained through cloud services and accompanying software packages. For example, on Amazon Web Services one can use the ParallelCluster software to launch a SLURM cluster.

Depending on the used hardware and available disk space, certain experiments may not be feasible. We have made an effort to reduce the memory requirements for running performance experiments on the largest data sets, and we expect that GPUs with 16GB of memory and machines with 128GB of main memory will be sufficient for running all or almost all of the distributed experiments.

\subsubsection{Software dependencies}

Reasonably up-to-date NVIDIA drivers must be installed on the machine. For the distributed experiments, a SLURM cluster is required. The remaining external software dependencies can be resolved using the \texttt{conda} package manager.

\subsubsection{Data sets}

Graph data sets for node property prediction are taken from the Open Graph Benchmark (OGB) repository. To decrease the time and minimum hardware resources required for experiments, we have provided the option to download preprocessed versions of the graph data. If electing to not download the preprocessed graphs, the first execution of the code on a new graph will download it from OGB and perform pre-processing locally. Note that, for the distributed experiments, additional pre-processing is needed to generate partition labels and reorder vertices by partition; this can be achieved using provided scripts, as described in the artifact README.

\subsection{Installation}

We recommend referring to the installation instructions provided at \url{https://github.com/MITIBMxGraph/SALIENT_plusplus_artifact/blob/main/INSTALL.md}. We summarize the installation process here.

\textbf{Installation in Python environment:} We provide instructions to install the artifact in a Python virtual environment. The installation can be used for both the single-machine and distributed multi-GPU experiments. The instructions for installing the artifact in a \verb|conda| environment are summarized below.

{\scriptsize
\begin{verbatim}
# Download and install miniconda
wget https://repo.anaconda.com/miniconda/\
    Miniconda3-py38_4.10.3-Linux-x86_64.sh
bash Miniconda3-py38_4.10.3-Linux-x86_64.sh

# Create a conda environment for the artifact
conda create -n salientplus python=3.9.5 -y
conda activate salientplus

# Install Pytorch, PyG, OGB, prettytable
conda install -y pytorch==1.13.1 torchvision==0.14.1 \
  torchaudio==0.13.1 pytorch-cuda=11.7 -c pytorch -c nvidia
conda install -y -c conda-forge ogb
conda install -y pyg -c pyg -c conda-forge
conda install -y pytorch-sparse -c pyg
conda install -y -c conda-forge nvtx
conda install -y -c conda-forge matplotlib
conda install -y -c conda-forge prettytable

# Install fast_sampler
cd fast_sampler
python setup.py install
cd ..

# (Optional) Install METIS
# - omitted, see INSTALL.md in the artifact repository
\end{verbatim}
}

\subsection{Experiment workflow}

We recommend referring to the artifact evaluation documentation located in the GitHub repository at \url{https://github.com/MITIBMxGraph/SALIENT_plusplus_artifact/blob/main/README.md}. We summarize the experimental workflow here. Unless otherwise noted, all file paths are relative to the \texttt{experiments} directory in the repository.

\subsubsection*{Initial setup for experiments}

We provide two utility scripts to streamline the process of downloading the pre-processed datasets and partitionings.  A configuration script (\verb|configure_for_environment.py|) determines which datasets to download based on the available disk space and the maximum number of GPUs (and partitions) to use for the experiments. A separate script (\verb|download_datasets_fast.py|) downloads the datasets selected during configuration. Please see the artifact repository README for instructions on using these setup scripts.

\subsubsection*{Simulation experiments}  

The script \verb|run_sim_experiments_paper.sh| runs simulation experiments on the ogbn-papers100M dataset to reproduce the results in \figref{cache-comm-vol-comparison-repfactor}. This script assumes that the 8-partition ogbn-papers100M dataset was downloaded using the setup scripts described above.

The driver script for running custom simulation experiments is \verb|run_sim_experiments.sh|.  The custom simulation script options are documented in the artifact README.

\subsubsection*{Distributed multi-GPU experiments}

These experiments require the use of a SLURM cluster with GPU nodes or a local machine with multiple GPUs. Each of the distributed-experiment scripts provides options to customize the experiment, and will display a table of results in the terminal after the experiment has completed. 

A summary of the provided scripts is as follows:

\begin{itemize}[noitemsep,topsep=0pt,parsep=0pt,partopsep=0pt]
\item \verb|experiment_optimization_impact.py| reproduces the experiment in \tabref{performance-progression} and \figref{ablation-study} to show the end-to-end impact of different optimizations on per-epoch runtime.
\item \verb|experiment_vary_replication_factor.py| reproduces the experiments in \cref{fig:scalability,fig:replication-factor-impact} to show the performance scalability with increasing number of distributed nodes and the relationship between per-epoch runtime and replication factor.
\item \verb|experiment_vary_gpu_percent.py| reproduces the experiment in \figref{cpu-vs-gpu-storage-impact} to analyze the performance impact of storing features on GPU.
\item \verb|experiment_accuracy.py| performs end-to-end training on the datasets and reports accuracy on the validation and test sets. 
\end{itemize}

If running these scripts on a SLURM cluster, it is necessary to configure the provided experimental driver script \verb|exp_driver.py| by editing the \verb|SLURM_CONFIG| variable at the top of the file. Additional details are provided in the artifact evaluation guide in the repository. If running these scripts locally on a single machine with multiple GPUs, one must pass the command line argument \verb|--run_local 1| to the script.

Users who wish to run custom experiments may use the core experiment driver script \verb|exp_driver.py| directly.

\subsection{Evaluation and expected result}

Upon completion of the simulation experiments, a table will be produced with the GNN training communication volume for different caching policies. This reproduces the data shown in \figref{cache-comm-vol-comparison-repfactor}.

Upon completion of the distributed GPU experiments, multiple tables will be produced with the results of \tabref{performance-progression} and Figures 4--7. Participants might opt to execute these experiments on different datasets or for different parameters depending on their hardware and time constraints.

\subsection{Experiment customization}

The following experiment customizations are possible. The simulation and distributed experiment scripts provide command-line options to run on different datasets and with different parameters. The experimental driver script \verb|exp_driver.py| may be used directly to run custom performance experiments in the multi-machine multi-GPU setting on different datasets, using different model architectures, and training parameters. The software may be used separately from the experimental driver scripts. The included pre-processing utility scripts can be used to partition graphs and reorder datasets according to a graph partitioning.

%
%
%

\section{Code Repository}
In addition to the artifact repository that focuses on benchmarking and reproducibility, an implementation of \oursystem
 for general-purpose usage is available at \url{https://github.com/MITIBMxGraph/SALIENT_plusplus}.


\section{Proof for layer-wise probabilities in \cref{prop:prob-exp-nhood}}
\label{app:proof-prob-exp-nhood-exact}

We start by considering the probability that some vertex $u \in \vertices$ is
sampled as a 1-hop neighbor of some minibatch $\batch$.  This is equal to the
probability that, for any vertex $v$, all of the following are true: $v$ is
included in the minibatch, $u$ is a neighbor of $v$, and $u$ is sampled among
$v$'s direct neighbors.  If we denote by $\nhoodsampled\oflayer[\hop](\batch)$ the vertices in the expanded neighborhood that are sampled after exactly $\hop$ expansion steps away from $\batch$, we have:
\begin{align*}
  p\ofhop[1](u)
  &= \Prob{u \in \nhoodsampled\oflayer[1](\batch)}
  \\
  &= \Prob{\bigunion_{v \in \nhood^{\tp}_{1}(u)} \Big(
    (v \in \batch) \intersection
    \left( u \in \nhoodsampled_{1}(v) \right)
    \Big)}
  \\
  &= 1 - \Prob{\bigintersection_{v \in \nhood^{\tp}_{1}(u)} \Big(
    (v \notin \batch) \union
    \left(u \notin \nhoodsampled_{1}(v)\right)
    \Big)}
  \\
  &= 1 - \prod_{v \in \nhood^{\tp}_{1}(u)}
    \Prob{(v \notin \batch) \union (u \notin \nhoodsampled_{1}(v))}
  \\
  &= 1 - \prod_{v \in \nhood^{\tp}_{1}(u)} \Big( \Prob{v \notin \batch} \Big.
  \\ &\hspace*{7em}
       \Big.
       + \Prob{(u \notin \nhoodsampled(v)) \intersection (v \in \batch)}
       \Big)
  \\
  &= 1 - \prod_{v \in \nhood^{\tp}_{1}(u)} \Big(
    1 - p\ofhop[0](v) + \big( 1 - \weight\oflayer[1](u,v) \big) \, p\ofhop[0](v)
    \Big)
  \\
  &= 1 - \prod_{v \in \nhood^{\tp}_{1}(u)} \Big(
    1 - \weight\oflayer[1](u,v) \, p\ofhop[0](v)
    \Big) .
\end{align*}
The product in the 4th line follows from the assumption of neighborhood-sampling independence between vertices in the minibatch.   Each
factor therein is the probability that $u$ is not sampled as a neighbor of $v$.
For directed graphs, when we write $\nhood^{\tp}_{1}(u)$ for the union,
intersection, and product indices, we mean the outgoing direct neighbors of
$u$.  For undirected graphs, $\nhood^{\tp}_{1}(u) = \nhood_{1}(u)$.

By a similar reasoning, the probability that a vertex $u$ is sampled
exactly $\layer$ hops away when expanding the neighborhood of some minibatch
is
\begin{align*}
  p\ofhop[\hop](u)
  &= \Prob{u \in \nhoodsampled\oflayer[\hop](\batch)}
  \\
  &= \Prob{\bigunion_{v \in \nhood^{\tp}_{1}(u)} \Big(
    (v \in \nhoodsampled\oflayer[\hop-1](\batch)) \intersection
    \left( u \in \nhoodsampled\oflayer[1](v) \right)
    \Big)}
  \\
  &= \cdots
  \\
  &= 1 - \prod_{v \in \nhood^{\tp}_{1}(u)} \Big(
    1 - \weight\oflayer[\hop](u,v) \, p\ofhop[\hop-1](v)
    \Big) ,
\end{align*}
where the omitted steps are similar to the ones for $p\ofhop[0](u)$ above.

\section{Walkthrough of \oursystem's pipelining stages}
\seclabel{pipeline-stages}
The minibatch preparation process proceeds in the following stages. Stage 1 obtains the next processed minibatch by \oursystem from
the neighborhood sampler.
Stage 2 performs an all-to-all communication to broadcast the number of remote vertices each machine will send/receive. Stage 3 transfers the metadata from stage 2 to the CPU so that appropriately sized tensors can be allocated. Stage 4 performs an all-to-all communication in which each machine $i$ sends to each machine $j$ a list of nodes whose features are local to machine $j$ and needed by machine $i$. Stage 5 receives the list of nodes requested by other machines, maps their global identifiers to local identifiers, and performs a device-to-host transfer so that the list of requested nodes is accessibly from CPU memory. Stage 6 launches an asynchronous worker thread that performs a ``masked selection'' operation to split each list of requested node indices into two groups based on whether the referenced node is stored in the local CPU or GPU memory. The background thread launched by Stage 6 also performs tensor slicing for the requested nodes whose features are stored in CPU memory. Note that the, seemingly innocuous, operation of performing a ``masked selection'' operator to divide indices into a CPU group and a GPU group can induce a device synchronization if it is performed on the GPU in Stage 5, which is why we assign this task to the background CPU thread. Stage 7 starts a host-to-device data transfer to send the results of Stage 6 to the GPU. Stage 8 performs tensor slicing on GPU for requested node features stored locally on GPU, and combines the CPU/GPU results so that features requested by each remote machine are in a single tensor. Additionally,
Stage 8 slices the machine's local feature cache to extract the remote features needed by the current machine~\footnote{The placement of these operations in the pipeline is somewhat arbitrary, as these operations do not depend on non-local information.}  Stage 9 performs an all-to-all communication to communicate all requested remote features. Stage 10 combines the received remote features into a single tensor that is permuted so as to be consistent with the locally-generated message-flow graph for the batch.

\end{document}